\documentclass{article}

\usepackage{microtype}
\usepackage{graphicx}
\usepackage{subcaption}
\usepackage{booktabs} 

\usepackage{hyperref}

\usepackage[preprint]{icml2026}

\usepackage{amsmath}
\usepackage{amssymb}
\usepackage{mathtools}
\usepackage{amsthm}

\usepackage{xcolor}
\usepackage{bm}
\graphicspath{{figures/}}     

\usepackage[table]{xcolor}
\usepackage{enumitem}

\usepackage[capitalize,noabbrev]{cleveref}

\theoremstyle{plain}

\theoremstyle{definition}

\theoremstyle{remark}

\usepackage[textsize=tiny]{todonotes}

\icmltitlerunning{Least but not last}

\begin{document}

\title{}

\twocolumn[
  \icmltitle{Least but not Last: Fine-tuning Intermediate Principal Components \\ for Better Performance-Forgetting Trade-Offs}



  \icmlsetsymbol{equal}{*}

  \begin{icmlauthorlist}
    \icmlauthor{Alessio Quercia}{ias8,rwth}
    \icmlauthor{Arya Bangun}{ias8}
    \icmlauthor{Ira Assent}{ias8,aarhus}
    \icmlauthor{Hanno Scharr}{ias8}
  \end{icmlauthorlist}

  \icmlaffiliation{rwth}{Department of Computer Science, RWTH Aachen University, Aachen, Germany}
  \icmlaffiliation{ias8}{IAS-8, Forschungszentrum Juelich, Juelich, Germany}
  \icmlaffiliation{aarhus}{Aarhus University, Aarhus, Denmark}

  \icmlcorrespondingauthor{Alessio Quercia}{a.quercia@fz-juelich.de}

  \icmlkeywords{Transfer learning, Parameter-efficient fine-tuning, Low-rank adaptation, Catastrophic forgetting, Continual learning, Large language models}

  \vskip 0.3in
]



\printAffiliationsAndNotice{}  

\begin{abstract}
Low-Rank Adaptation (LoRA) methods have emerged as crucial techniques for adapting large pre-trained models to downstream tasks under computational and memory constraints. 
However, they face a fundamental challenge in balancing task-specific performance gains against catastrophic forgetting of pre-trained knowledge, where existing methods provide inconsistent recommendations. 
This paper presents a comprehensive analysis of the performance-forgetting trade-offs inherent in low-rank adaptation using principal components of weight matrices as initialization.
Our investigation reveals that fine-tuning intermediate components leads to better balance and robustness to high learning rates than first (PiSSA) and last (MiLoRA) components in existing work.
%
Building on these findings, we provide practical guidelines for initialization of LoRA methods to balance the performance-forgetting trade-off. In a thorough empirical study on a variety of computer vision and NLP tasks we confirm that these guidelines achieve high accuracy and reduced forgetting.
\end{abstract}

\begin{figure}
    \centering
    \includegraphics[width=0.8\columnwidth]{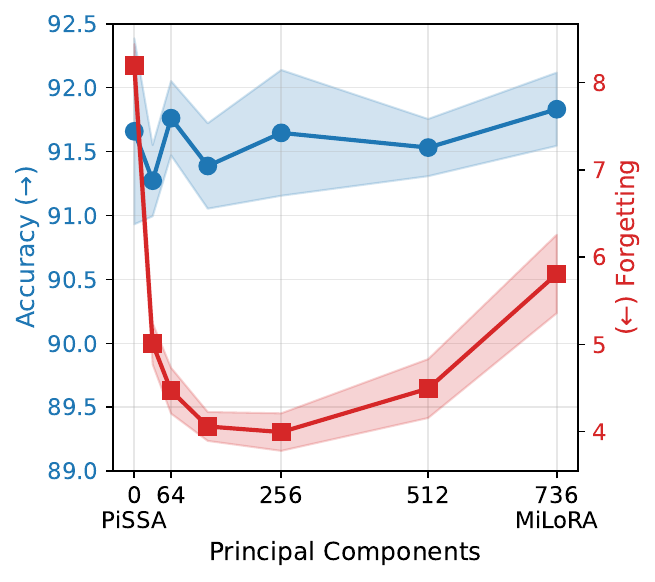}
    \caption{Accuracy (left) and forgetting (right) when fine-tuning principal components on ImageNet1k pre-trained ViT-B to Caltech101. Forgetting shows a U-shape with most information lost at the extremes where existing methods PiSSA use the main, and MiLoRA the least components, respectively. }
    \label{fig:caltech101_highlight}
\end{figure}

\section{Introduction}

The explosive growth of large-scale pre-trained models has revolutionized artificial intelligence across multiple domains, from Natural Language Processing to Computer Vision. Foundation models, often containing billions or even trillions of parameters, demonstrate remarkable capabilities in learning and in generating human-like content.
However, their deployment in real-world applications faces significant computational and memory constraints that present considerable challenges, in particular in continual learning scenarios, which require constant knowledge updates.


Traditional fine-tuning approaches require updating of all model parameters, leading to substantial computational and memory requirements, which can be prohibitive for many applications \citep{quercia2025data}.
For instance, fine-tuning a large language model with billions of parameters demands enormous GPU memory and computational resources, making it inaccessible to institutions with limited compute resources or financial budget. This bottleneck is compounded when models need to be adapted to multiple downstream tasks or domains, as each adaptation traditionally necessitates a complete copy of the fine-tuned model.

Parameter-Efficient Fine-Tuning (PEFT) methods have emerged as a promising solution, enabling effective adaptation of large models with significantly reduced resource requirements \citep{hu2021lora,liu2024dora,kopiczko2024vera,quercia20251lora,meng2024pissa,wang2024milora,zaken2022bitfit,xie2023difffit}. Among the most notable approaches, Low-Rank Adaptation (LoRA) \citep{hu2021lora} constrains weight updates to low-rank matrices, which has been shown to preserve much of the pre-trained knowledge while requiring only a small fraction of the parameters for tuning. Notably, as \citep{biderman2024lora} states ``\textit{LoRA learns less and forgets less}'', indicating LoRA's advantage in retaining pre-trained capabilities, especially important in continual learning applications.

LoRA has attracted substantial attention in the research community and inspired numerous extensions, including DoRA \citep{liu2024dora}, which decomposes weights into magnitude and direction components, while VeRA, \citep{kopiczko2024vera}, introduces a vector-based random matrix adaptation for further parameter reduction.  More recent advancements include PiSSA \citep{meng2024pissa} and MiLoRA \citep{wang2024milora}, which respectively initialize LoRA modules with essential principal components for better adaptation, and minor principal components to reduce forgetting. Other PEFT methods study alternative approaches like BitFit \citep{zaken2022bitfit}, which only updates bias terms, and DiffFit \citep{xie2023difffit}, which specializes in diffusion model adaptation through targeted adjustments to scaling factors and biases.

\begin{figure}
    \centering
    \includegraphics[trim={0 0.5cm 4cm 0},clip,width=0.5\textwidth]{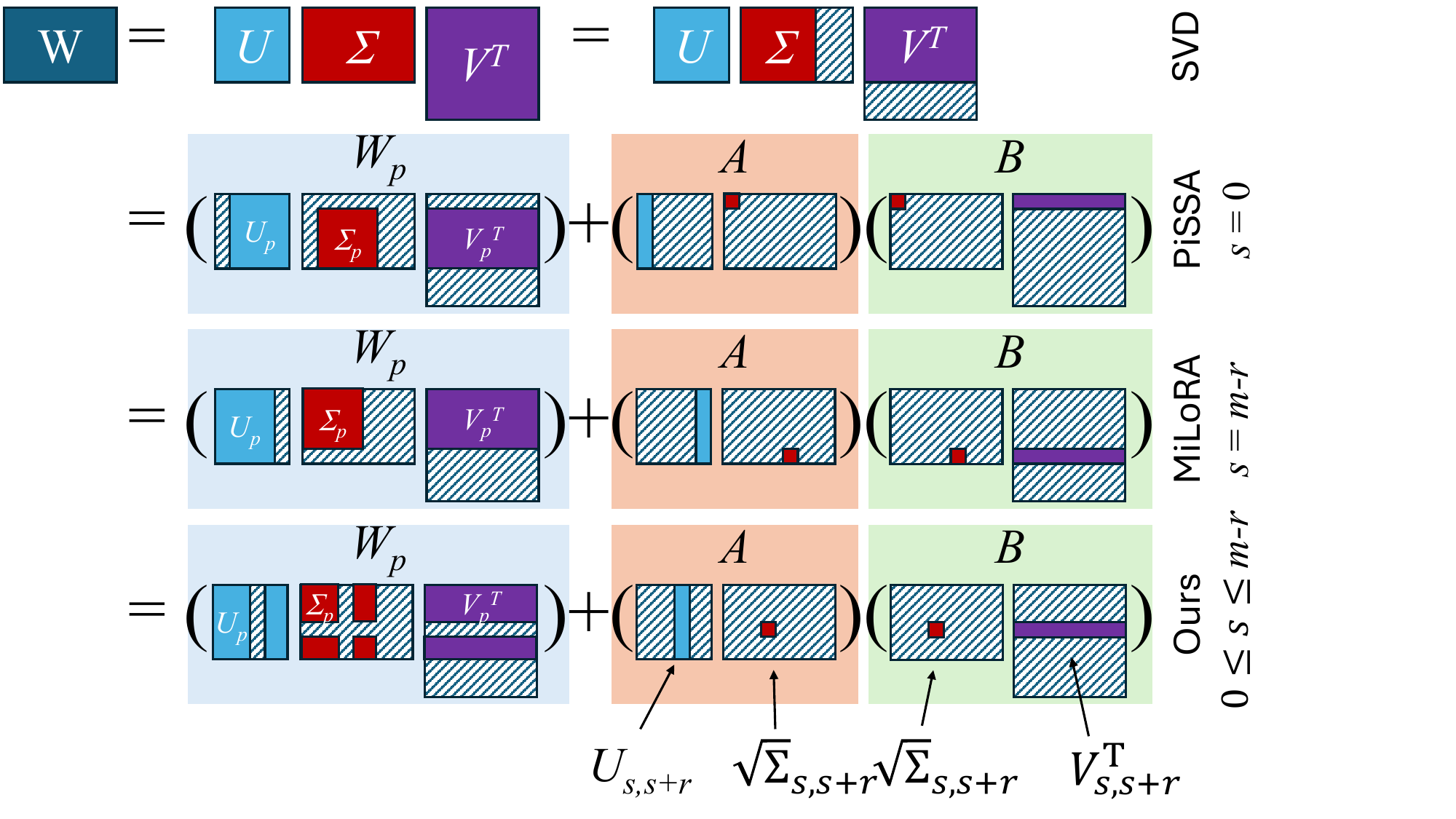}
    \caption{PiSSA, MiLoRA and our proposed approach.}
    \label{fig:caltech101_methods}
\end{figure}

Despite their efficiency, PEFT methods face a challenge that has been relatively understudied: the \textit{performance-forgetting trade-off}. While PEFT approaches excel at task adaptation with minimal resource investment, they risk sacrificing original pre-trained knowledge.
While LoRA emphasizes the importance of learning more and forgetting less \citep{biderman2024lora}, the precise relationship between rank, adaptation capacity, and catastrophic forgetting remains an open question. 

Recent studies introduce the idea of using principal components to initialize LoRA modules \citep{meng2024pissa,wang2024milora}, where MiLoRA explicitly targets the performance-forgetting trade-off problem, arguing that PiSSA forgets more as it fine-tunes the main principal components of a pre-trained model, which contain the main information, whereas last components contain long-tail or complementary information \citep{wang2024milora}. Differently from PiSSA and MiLoRA, we investigate intermediate components, under the assumption that both main and long-tail information needs to be preserved, especially in image classification applications. As we show in Figure~\ref{fig:caltech101_highlight}, this assumption is confirmed by our study, where we see a U-shaped forgetting curve spanning over the principal components, and suggesting that fine-tuning intermediate components leads to reduced forgetting.

Understanding and addressing these trade-offs is crucial, especially for deploying foundation models in continual learning \citep{wang2024comprehensive,verwimp2023continual}, or multi-task \citep{vandenhende2021multi,zhang2021survey,quercia2025enhancing} settings where the ability to adapt without sacrificing pre-trained knowledge is vital.


Our contributions are summarized as follows:
\begin{itemize}[itemsep=0pt, topsep=0pt, parsep=0pt]
    \item We propose a generalization of PiSSA and MiLoRA, enabling contiguous component selection for fine-tuning. 
    \item We offer a systematic analysis of performance-forgetting trade-offs across a variety of LoRA variants, across many image classification and NLP datasets.
    \item Across our experiments, neither PiSSA nor MiLoRA yields the most favorable performance–forgetting trade-offs; instead, varying the fine-tuned components from first to last produces a U-shaped forgetting curve while accuracy saturates to a similar maximum, indicating that intermediate components offer the best balance between accuracy and retention.
    \item By analyzing principal components before and after fine-tuning, we find that the U-shape arises from unintended damage to unfocused components, with particularly severe degradation when early components are perturbed.
\end{itemize}

\section{Related Work}
The rapid scaling of foundation models with billions of parameters has 
transformed AI capabilities across Natural Language Processing and Computer 
Vision.
However, full fine-tuning 
of these models remains computationally expensive, requiring extensive GPU 
memory and storage for each task adaptation.

\textbf{Parameter-Efficient Fine-Tuning (PEFT)} addresses this challenge by 
updating only a small fraction of parameters. LoRA~\citep{hu2021lora} 
constrains weight updates to low-rank matrices, achieving comparable performance 
to full fine-tuning with a smaller percentage of parameters while preserving more of the pre-trained knowledge. Biderman et al.\ \citep{biderman2024lora} note that ``LoRA learns less and forgets less,'' highlighting its advantages in knowledge preservation, yet subsequent variants like DoRA~\citep{liu2024dora} enhance adaptation by decomposing weights into magnitude and direction components. Similarly, PiSSA~\citep{meng2024pissa} and MiLoRA~\citep{wang2024milora} propose to initialize LoRA modules using main and minor principal components of the target pre-trained model, respectively, for enhanced efficiency on one hand, and for knowledge preservation on the other hand. On the other hand, OLoRA~\citep{buyukakyuz2024olora} uses orthonormal matrix initialization via QR decomposition to improve adaptation. In parallel, variants like VeRA~\citep{kopiczko2024vera} and 1LoRA~\citep{quercia20251lora} further minimize trainable parameters, with VeRA employing shared random matrices and layer-specific scaling vectors, and 1LoRA consolidating to a single trainable vector per module. Other LoRA variants dynamically prune less important ranks during training (AdaLoRA \citep{zhang2023adalora}), or combine LoRA with 4-bit quantization for memory savings (QLoRA \citep{dettmers2023qlora}). Lastly, other PEFT methods include BitFit~\citep{zaken2022bitfit}, which achieves extreme efficiency by tuning only bias terms, and DiffFit~\citep{xie2023difffit}, which specializes in diffusion model adaptation.

In this work, propose a generalization of PiSSA and MiLoRA allowing for fine-grained rank selection for fine-tuning. We compare this generalization to LoRA~\citep{hu2021lora} and LoRA decomposition-based initialization methods \citep{meng2024pissa,wang2024milora,buyukakyuz2024olora,liu2024dora} with fixed rank and highlight their performance-forgetting trade-offs. We therefore exclude dynamic rank selection methods like AdaLoRA \cite{zhang2023adalora} for comparability of fixed rank methods.

\section{Learning and forgetting}
Pre-training large models requires prohibitive computational costs—often millions of GPU hours and billions in infrastructure. Consequently, adapting foundational models has emerged as the practical strategy for downstream specialization. Given this reality, along with impending data scarcity where most available data has already been consumed by prior models, fine-tuning of pre-trained models on newly available data while preserving prior knowledge will become increasingly critical.

Recent approaches attempt to address this performance-forgetting trade-off through targeted low-rank updates \citep{hu2021lora}: these methods constrain fine-tuning updates to a low-dimensional subspace by factorizing weight changes as the product of two low-rank matrices $A$ and $B$ as $\Delta W = AB \quad (A\in\mathbb{R}^{m\times r},\, B\in\mathbb{R}^{r\times n},\, r\ll\min(m,n))$, drastically reducing the number of trainable parameters while preserving expressive power. Rather than updating all model weights directly, only these compact low-rank factors are optimized, enabling efficient adaptation with minimal interference to pre-trained knowledge.

In this paper, we consider LoRA and LoRA initialization methods with fixed rank, and in particular we focus on principal-component-based initialization methods like PiSSA \citep{meng2024pissa} and MiLoRA \citep{wang2024milora}, studying their differences and proposing a better alternative in terms of performance-forgetting trade-offs.

\textbf{Catastrophic forgetting} remains a critical limitation. While PEFT 
methods reduce resource demands, they often exhibit performance-forgetting 
trade-offs, as highlighted in \cite{wang2024milora}.
Adaptive rank allocation and regularization 
strategies \citep{zhang2023adalora} show promise dynamically adjusting low-rank dimensions to balance expressivity and stability and penalizing deviations in critical subspaces, but systematic 
comparisons for fixed rank principal components-based LoRA initialization methods are missing. In particular in-depth analyses of components and their effect on performance-forgetting have not yet been provided for fixed rank LoRA methods.

\subsection{Choosing principal components}
Unlike prior work focusing on isolated methods, we provide comprehensive 
analysis of the performance-forgetting dynamics across LoRA variants of fixed rank, quantifying stability-plasticity trade-offs to guide robust 
LoRA initialization design.

PiSSA~\citep{meng2024pissa} leverages the intuition that the largest principal components of weight matrices capture the most expressive directions for new task performance. By selectively fine-tuning only these high-magnitude singular directions—while leaving smaller components frozen—it maximizes downstream adaptation capacity without broadly disrupting the model's geometry. Conversely, MiLoRA~\citep{wang2024milora} exploits the hypothesis that the smallest principal components represent task-orthogonal subspaces minimally utilized by prior training. Targeting these low-magnitude directions for fine-tuning minimizes interference with pre-trained representations while still providing sufficient expressivity for new tasks, achieving a principled forgetting-performance trade-off through spectral separation.
Ideally, methods should optimize new task accuracy while preserving prior knowledge; however, we empirically demonstrate several fine-tuning scenarios where neither PiSSA nor MiLoRA achieve optimal performance-forgetting trade-offs. 
For this reason, we propose to fine-tune intermediate principal components, rather than the extremes.

We summarize PiSSA and MiLoRA in Figure~\ref{fig:caltech101_methods}, highlighting our proposed  generalization that recommends fine-tuning intermediate components, based on our analysis and empirical findings.

\subsection{Approach}
Let $W$ be a pre-trained $m \times n$ matrix, and $W = U \Sigma V^T$ its Singular Value Decomposition  (SVD), where $U$ and $\Sigma$ are $m \times m$ and $V^T$ is $m \times n$. We denote the considered matrix slices as $U_{s, s+r}$, $\Sigma_{s, s+r}$ and $V^T_{s, s+r}$, where $s$ is the starting component and $r$ is the rank. Therefore the decomposed matrix can be represented as $U = U_p + U_{s, s+r}$, $\Sigma = \Sigma_p + \Sigma_{s, s+r}$ and $V = V_p + V_{s, s+r}$. For example, the diagonal matrix is the sum of the following matrices
\begin{equation*}
    \begin{aligned}
        &\Sigma_{s,s+r} = diag(0, \ldots, 0, \sigma_{s}, \ldots, \sigma_{s+r-1}, 0, \ldots, 0) \\[1em]
        &\Sigma_{p} = diag(\sigma_{0}, \ldots, \sigma_{s-1}, 0, \ldots, 0, \sigma_{s+r}, \ldots, \sigma_{m})
    \end{aligned}
\end{equation*}

We can then define the LoRA \citep{hu2021lora} matrices as 
\begin{equation}
    A=U_{s,s+r}\Sigma^{1/2}_{s,s+r}
\quad \text{and} \quad  
    B=\Sigma^{1/2}_{s,s+r}V^T_{s,s+r}
\end{equation} 

and the forward pass as 
\begin{equation}
    Y = X(W_p + \Delta W) = X(W_p + AB)
\end{equation}
where $X$ represents the input dataset and $W_p = W - U_{s, s+r} \Sigma_{s, s+r} V^T_{s, s+r}$ be the residual pre-trained matrix, which will be frozen during fine-tuning.

Note that this is a generalization of PiSSA \citep{meng2024pissa} and MiLoRA \citep{wang2024milora}, where the former can be achieved with $s = 0$ and the latter with $s = m-r$.

\subsection{Component analysis}
\label{sec:analysis}
We present an in-depth analysis investigating why extreme principal components exhibit higher susceptibility to catastrophic forgetting under extended fine-tuning or high learning rates. We further show that prolonged training exacerbates this phenomenon. Thus, we derive the conditions under which extreme principal components (at both spectrum ends, corresponding to PiSSA \citep{meng2024pissa} and MiLoRA \citep{wang2024milora}) undergo destabilizing dynamics during sequential task adaptation. Our analysis reveals how fine-tuning principal components at the extremes leads to higher damage to the main singular values, confirmed empirically as non-monotonic forgetting behavior across the singular value spectrum (Section \ref{sec:exp}). We first examine these dynamics in parameter space before analyzing their implications in feature space.

\begin{figure}[t]
    \centering
    \includegraphics[width=0.40\textwidth]{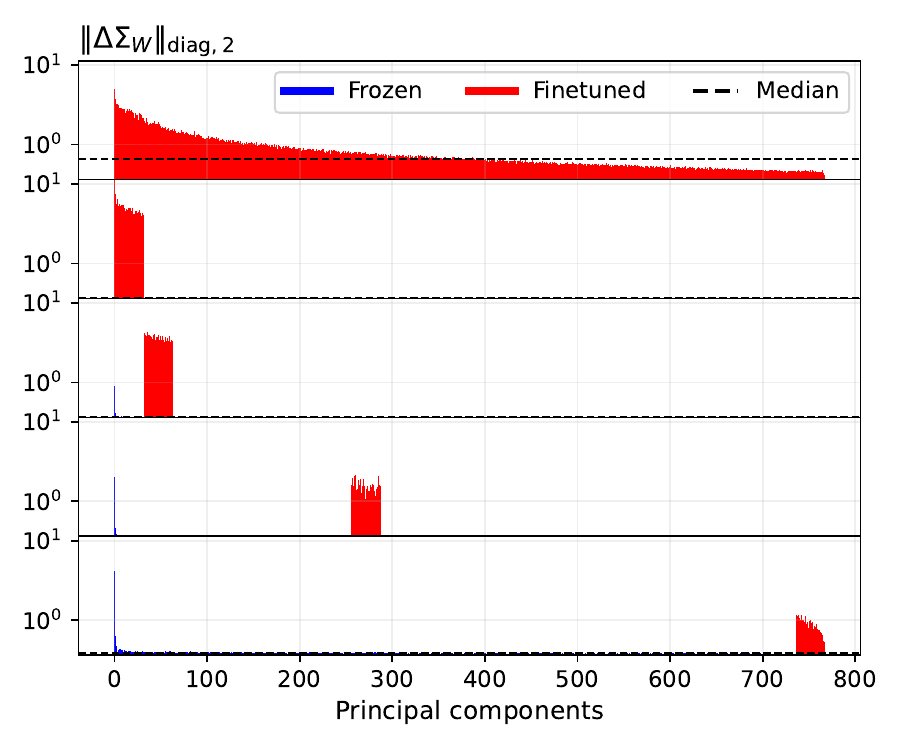}
%
    \centering
    \includegraphics[width=0.40\textwidth]{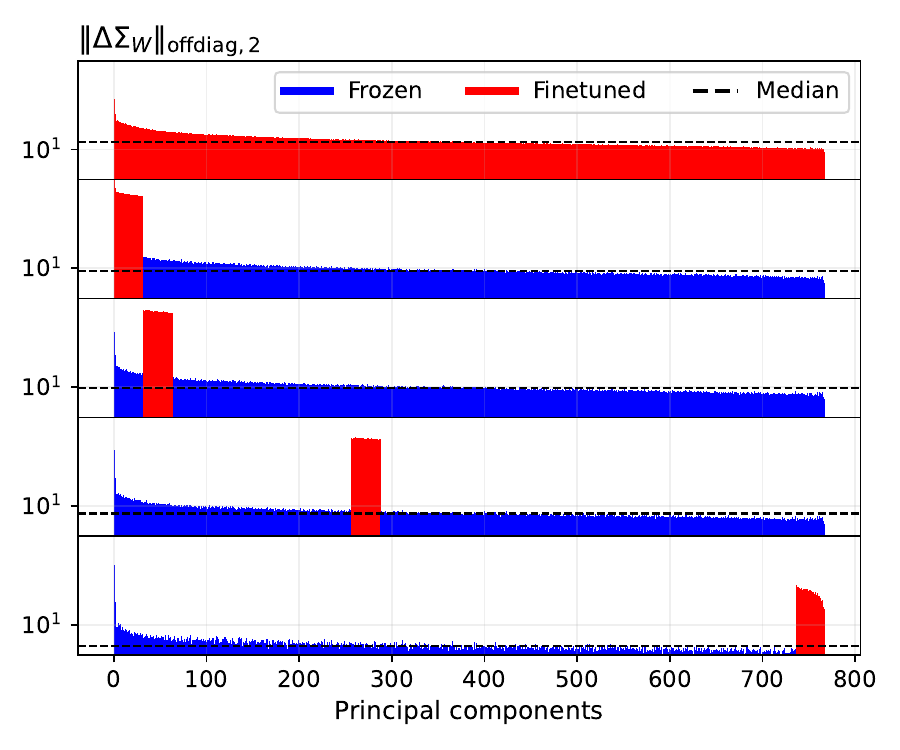}
    \caption{(ImageNet1k $\rightarrow$ Caltech101) Changes to the diagonal (top) and off-diagonal (bottom) in parameter space, $\operatorname{offdiag}(\Delta \Sigma_W)$, see Eq.\ \ref{eq:delta_offdiag}. We show the element-wise norm and column-wise norm $\| \operatorname{offdiag}(\Delta \Sigma_W)_{i.}\|_{2}$, respectively.}
    \label{fig:in_caltech101_params_diagoffdiag}
\end{figure}
\begin{figure}[t]
    \centering
    \includegraphics[width=0.30\textwidth]{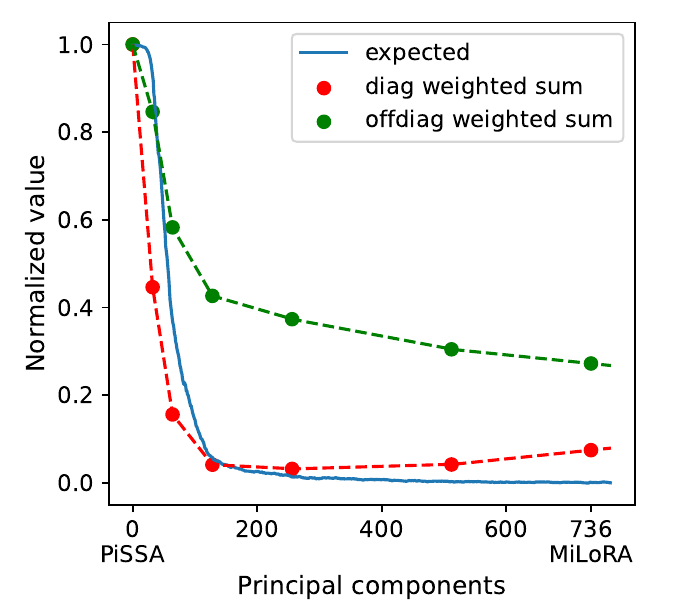}
    \caption{Normalized expected forgetting, scaled diagonal (red) and off-diagonal (green) sum of changes in parameter space.}
    \label{fig:in_caltech101_params_sum}
\end{figure}

\subsubsection{Parameter space}
Let $W_0$ and $W_{s,s+r}$ be the weight matrices of the considered pre-trained model before and after fine-tuning of components between $s$ and $s+r$. To study the forgetting behavior depending on which components are fine-tuned, we analyze the changes to the parameters after fine-tuning in principal component space. In practice, we compute the SVD of the original weight matrix $W_0$ 
\begin{equation}
    W_0 = U_{W_0} \Sigma_{W_0} V_{W_0}^T
\end{equation}
and then we project the fine-tuned weight matrix $W_{s,s+r}$ into its coordinate system 
\begin{equation}
    \Sigma_{W_{s,s+r}} = U_{W_0}^T W_{s,s+r} V_{W_0}
\end{equation}
where $\Sigma_{W_{s,s+r}}$ corresponds to $\Sigma_{W_0}$ after fine-tuning. We denote the changes as
\begin{equation}
    \Delta \Sigma_W = | \Sigma_{W_0} - \Sigma_{W_{s,s+r}} |.
\end{equation} 
We denote the diagonal and off-diagonal of $\Delta \Sigma_W$ by 
\begin{equation}
    \operatorname{diag}(\Delta \Sigma_W) = (\Delta \Sigma_{W11},\dots,\Delta \Sigma_{Wnn})
    \label{eq:delta_diag}
\end{equation}
and 
\begin{equation}
\begin{aligned}
    &\operatorname{offdiag}(\Delta \Sigma_W) = \Delta \Sigma_W - \operatorname{diag}(\Delta \Sigma_W) \\
    &= (\Delta \Sigma_{Wij})_{i,j=1}^n \quad \text{with $i\neq j$}.
\end{aligned}
    \label{eq:delta_offdiag}
\end{equation}

respectively. By this, we disentangle the changes in principal values (Eq.\ \ref{eq:delta_diag}) from those in principal directions within the SVD of weight update matrices, and relate this to the observed empirical behavior in the experiments (Sect.~\ref{sec:exp}). This decomposition isolates the scaling of principal components from their directional shifts (off-diagonal changes), enabling precise attribution of behavioral changes to specific subspaces of the parameter space. By analyzing these distinct contributions separately, our approach reveals how fine-tuning modifies the geometry of weight updates across different principal components. 
This framework facilitates a deeper analysis of forgetting phenomena by linking the observed U-shaped forgetting curve to the changes in principal components. 

In particular, we show the analysis corresponding to results in Figure~\ref{fig:caltech101_highlight}. Figure~\ref{fig:in_caltech101_params_diagoffdiag} shows the diagonal (top) and off-diagonal (bottom) L2 norms of the models in the considered experiment. From top to bottom, we display fine-tuning all parameters, fine-tuning components 0-32 (PiSSA), 32-64, 256-288, and 736-768 (MiLoRA).
We observe from the figures, that fine-tuning in a certain low-rank region changes the respective singular values most, as expected. However, also other 'frozen' components are changed due to rotation of the 'hot' low rank subspace. Specifically, it emerges that fine-tuning the extremes (PiSSA or MiLoRA) also leads to higher changes to the very first principal component, and subsequent ones, suggesting that main information from the previous task might be damaged more.

In Figure~\ref{fig:in_caltech101_params_sum} we show a summary of the previous figures, by computing the sum over the components, weighted by their `expected' contribution $p$ to forgetting. To generate $p_i$, we zero out the $i$-th principal component of the pretrained model, evaluate on the prior data, and compute the resulting forgetting value $f_i$. Finally, we normalize $p_i = f_i / \max_j(f_j)$. $p$ is shown as blue line. We then compute the weighted sums $\sum_i p_i \cdot \|\operatorname{diag}(\Delta \Sigma_W)_{ii}\|_{2}$ (red line) and $\sum_i p_i \cdot \| \operatorname{offdiag}(\Delta \Sigma_W)_{i.}\|_{2}$ (green line) for the diagonal and off-diagonal of $\Delta \Sigma_W$, respectively, see Eqs.\ \ref{eq:delta_diag} and \ref{eq:delta_offdiag}. The figure highlights that in contrast with the `expected' behaviour ($p$, blue line), where forgetting decreases monotonically with increasing principal components, the weighted diagonal (red line) forms a soft U-shape. This indicates that fine-tuning components at the extremes leads to higher damage in the pre-trained model. We show the changes in feature space in the next subsection.

\subsubsection{Feature space}
Let $W_0$ and $W_{s,s+r}$ be the weight matrices of the considered pre-trained model before and after fine-tuning of components between $s$ and $s+r$. And let $X_0$ be a small, random but fixed subset of the data used to pre-train $W_0$, and let
\begin{equation}
    Y_0 = X_0W_0
\end{equation}
be the outputs of each considered layer in model $W_0$ for inputs $X_0$. In this study we analyze the changes in feature space after fine-tuning. To do so, we compute the SVD
\begin{equation}
    Y_0 = U_{Y_0} \Sigma_{Y_0} V^T_{Y_0}    
\end{equation}
and then we project the outputs of fine-tuned weight matrix $Y_{s,s+r} = X_0W_{s,s+r}$ into its coordinate system as follows
\begin{equation}
    \Sigma_{Y_{s,s+r}} = U_{Y_0}^T Y_{s,s+r} V_{Y_0}
\end{equation}
where $\Sigma_{Y_{s,s+r}}$. We denote the changes in feature space as
\begin{equation}
    \Delta \Sigma_Y = | \Sigma_{Y_0} - \Sigma_{Y_{s,s+r}} |.
    \label{eq:delta_sigma_y}
\end{equation} 
Lastly, we denote the diagonal and off-diagonal of $\Delta \Sigma_Y$ as $\operatorname{diag}(\Delta \Sigma_Y)$ and $\operatorname{offdiag}(\Delta \Sigma_Y)$, respectively. Here, we disentangle the changes in principal values from those in principal directions, as done for the parameter space.

\begin{figure}[t]
    \centering
    \includegraphics[width=0.40\textwidth]{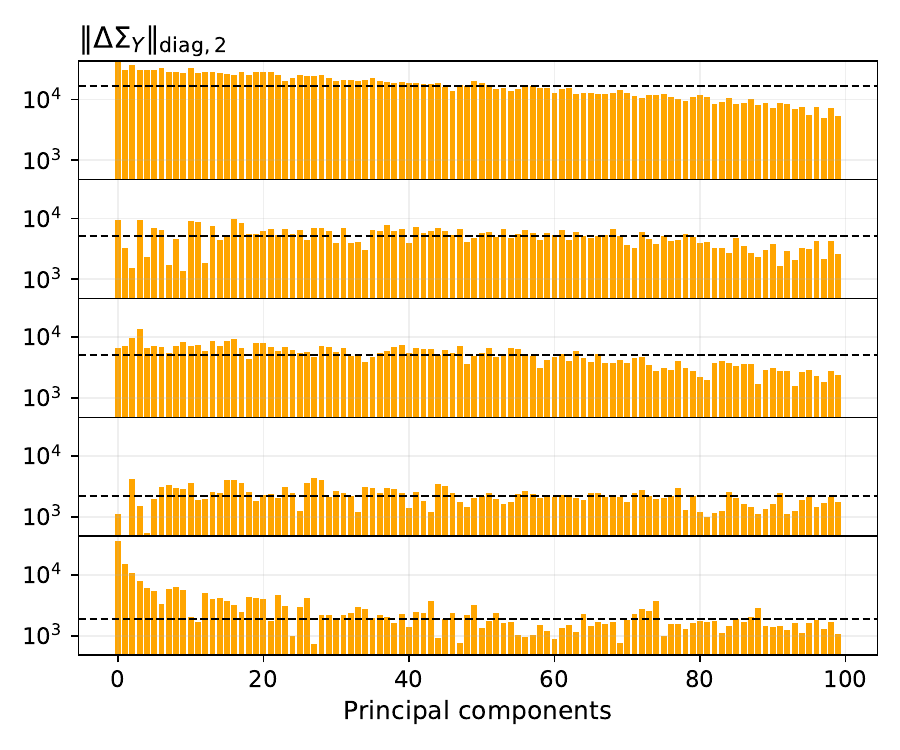}
%
    \centering
    \includegraphics[width=0.40\textwidth]{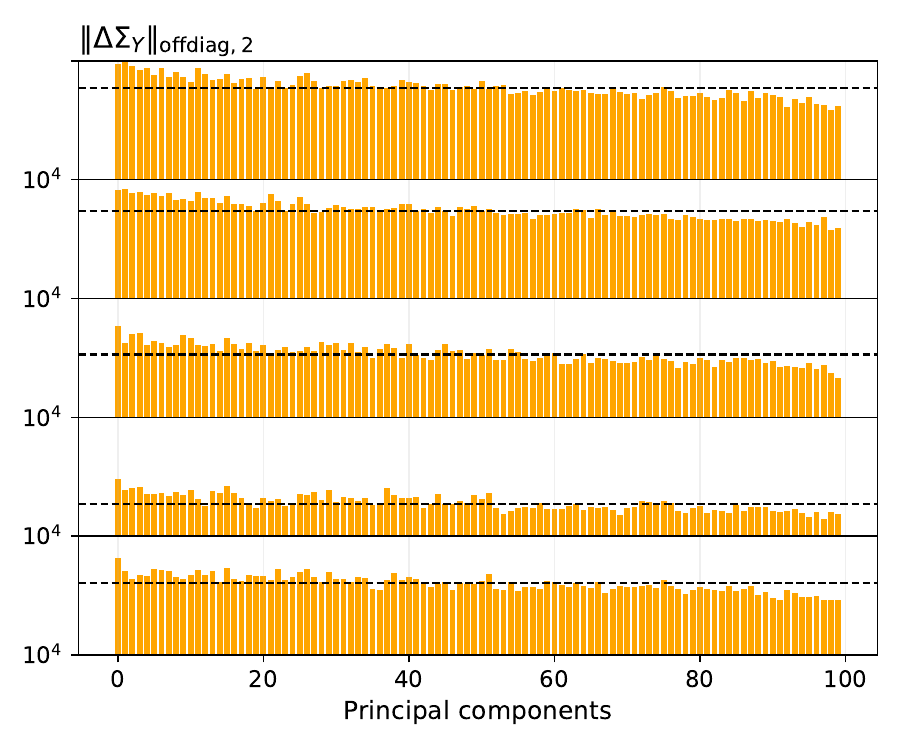}
    \caption{(ImageNet1k $\rightarrow$ Caltech101) Changes to the diagonal (top) and off-diagonal (bottom) in feature space, $\operatorname{diag}(\Delta \Sigma_Y)$ and $\operatorname{offdiag}(\Delta \Sigma_Y)$. We show the element-wise and column-wise norm $\| \operatorname{offdiag}(\Delta \Sigma_Y)_{i.}\|_{2}$, respectively.}
    \label{fig:in_caltech101_features_diagoffdiag}
\end{figure}
\begin{figure}[t]
    \centering
    \includegraphics[width=0.30\textwidth]{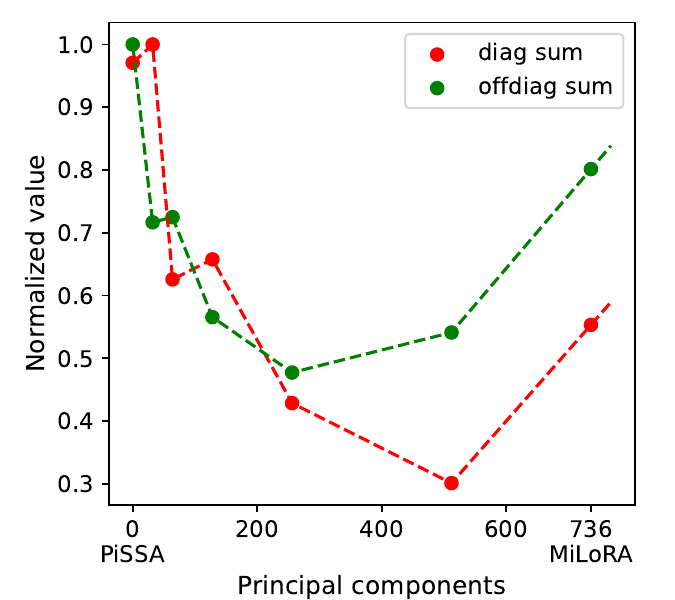}
    \caption{Normalized expected forgetting, scaled diagonal (red) and off-diagonal (green) sum of changes in feature space.}
    \label{fig:in_caltech101_features_sum}
\end{figure}

Figure~\ref{fig:in_caltech101_features_diagoffdiag} shows the changes in feature space for $X_0$ being a subset of 100 samples, displaying the diagonal (top) and off-diagonal (bottom) L2 norms of the models in the considered experiment. These figures confirm our previous analysis, showing that fine-tuning all parameters and extreme components (first 2 rows and last one) lead to higher changes in feature space, whereas fine-tuning intermediate components lead to less changes.

This is summarized in Figure~\ref{fig:in_caltech101_features_sum}, where we show the sum over the components, together with the `expected' distribution, where we see that changes in feature space form a U-shape, both in diagonal and off-diagonal. Please note that these analyses are made on single models, whereas in Figure~\ref{fig:caltech101_highlight} we report mean and standard deviation.

We observe, that the shallower U-shape in parameter space translates to a more pronounced U-shape in feature space.

\section{Experiments}
\label{sec:exp}
We conduct an extensive empirical study with two main focus points: (i) using our proposed analysis, we study the impact of the principal components used, and of the training time on forgetting and accuracy, and (ii) based on our findings, we propose a balanced trade-off method leveraging intermediate principal components. For the latter, we assess the effectiveness and robustness of the proposed method across both vision and language domains, comparing it against recent LoRA methods with fixed rank and similar number of parameters for comparability: LoRA \citep{hu2021lora}, DoRA \citep{liu2024dora}, OLoRA \citep{buyukakyuz2024olora}, PiSSA \citep{meng2024pissa}, MiLoRA \citep{wang2024milora}. Our study covers a broad spectrum of Image Classification cases, including models and datasets of varying scale, complexity, and number of classes. In addition, we systematically evaluate on diverse NLP tasks spanning mathematical reasoning, python coding and common sense tasks, thereby analyzing behavior under heterogeneous data distributions and task formats. 

    

\begin{figure*}[h]
    \centering 
    \includegraphics[width=0.5\textwidth, trim=0 11cm 0 0, clip]{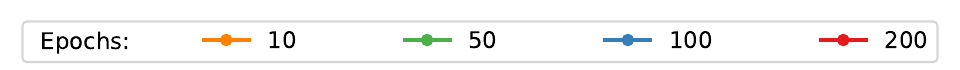} 
    
    \begin{subfigure}{0.30\textwidth}
        \centering
        \includegraphics[width=\textwidth]{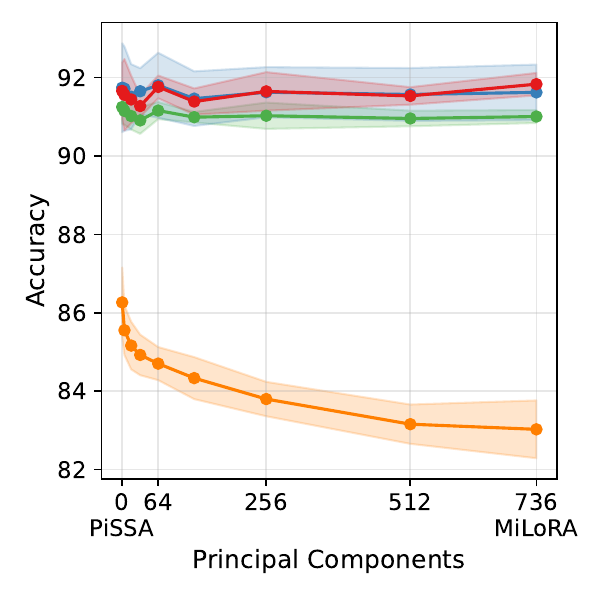}
    \end{subfigure}
    \hspace{2cm}
    \begin{subfigure}{0.30\textwidth}
        \centering
        \includegraphics[width=\textwidth]{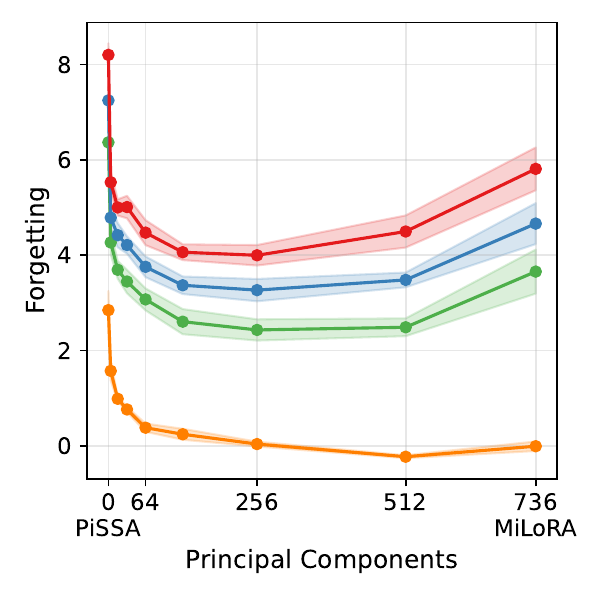}
    \end{subfigure}
    
    \caption{(ImageNet1k $\rightarrow$ Caltech101) Accuracy (left) and Forgetting (right) when fine-tuning an ImageNet1k pre-trained ViT-Base to Caltech101, using different principal component slices with starting points $s$ (x-axis) and rank 32.}
    \label{fig:in_caltech101}
\end{figure*}

\subsection{Image Classification}
We examine whether our findings generalize to other image classification datasets. We evaluate the impact on performance and forgetting of fine-tuning different principal component ranges, with starting points $s$ (0, 32, 64, 128, 256, 512, 736) with rank $r=32$, where $s=0$ is PiSSA \citep{meng2024pissa} and $s=736$ is MiLoRA \citep{wang2024milora}. We fine-tune an ImageNet1k pre-trained ViT-B on a variety of image classification datasets: CIFAR-10 \citep{krizhevsky2009learning}, CIFAR-100 \citep{krizhevsky2009learning}, DTD \citep{cimpoi2014describing}, Caltech-101 \citep{fei2007one}, Caltech-256 \citep{griffin2007caltech256}, Food-101 \citep{bossard2014food}, Oxford-IIIT Pets \citep{parkhi2012cats}, Oxford Flowers 102 \citep{nilsback2008automated}, Stanford Cars \citep{krause20133d}, Stanford Dogs \citep{khosla2011novel}, FGVC Aircraft \citep{maji2013fine}. In these experiments, we compute forgetting as absolute difference of accuracies (before and after fine-tuning). Note that this forgetting is correlated to the forgetting as computed in \citep{wang2024milora,kalajdzievski2024scaling}. Training details are reported in Appendix~\ref{app:cls_details}.

From Figure~\ref{fig:in_caltech101} we observe that forgetting exhibits a characteristic U-shaped curve when models are fine-tuned to high accuracy on a new task, i.e. long enough to fit it properly or to over-fit it. The longer the fine-tuning duration on the new task, the more pronounced this U-shape becomes. This pattern indicates that \textbf{both the highest and lowest principal components are particularly susceptible to catastrophic forgetting under extended fine-tuning}. Therefore the trade-off accuracy-forgetting is dominated by the forgetting, and the best value shows up ``in the middle", i.e. when fine-tuning intermediate principal components. This is shown in Appendix in Figures~\ref{fig:in_cifar10}--\ref{fig:in_dtd}.

We report a summary of our experiments in Table~\ref{tab:cls}, where the observed behaviour is confirmed, with more or less pronounced U-shapes. In summary, when a model is trained for long-enough, the accuracy on the new task seems to plateau and reach approximately the same value for each starting point, whereas the forgetting forms a U-shape, where extreme values are higher than intermediate ones, leading to intermediate components having better performance-forgetting trade-offs. This confirms the analysis of the U-shape phenomenon provided in Section~\ref{sec:analysis}. The Table additionally highlights that the range between 16\% and 66\% of the linear layer sizes is robustly leading to improved performance-forgetting trade-offs. In Table~\ref{tab:cls_all} we report the complete results with mean and standard deviation.


\begin{table*}[h]
    \setlength{\tabcolsep}{3pt} 
    \scriptsize
    \centering
    \caption{Accuracy and Forgetting of Imagenet1k and various finetuned classification datasets, after fine-tuning for up to 200 epochs. Maximum mean of 4 independent runs. We highlight $\textbf{best}$, and in \textcolor{yellow}{yellow} the best rows overall.}
    \begin{tabular}{ccccccccccccc}
    \toprule
    Methods & \multicolumn{2}{c}{CIFAR10} & \multicolumn{2}{c}{CIFAR100} & \multicolumn{2}{c}{DTD} & \multicolumn{2}{c}{Caltech101} & \multicolumn{2}{c}{Caltech256} & \multicolumn{2}{c}{Food101} \\
     & Acc\,$\!^{\scriptstyle \uparrow}$ & Forg\,$\!^{\scriptstyle \downarrow}$ & Acc\,$\!^{\scriptstyle \uparrow}$ & Forg\,$\!^{\scriptstyle \downarrow}$ & Acc\,$\!^{\scriptstyle \uparrow}$ & Forg\,$\!^{\scriptstyle \downarrow}$ & Acc\,$\!^{\scriptstyle \uparrow}$ & Forg\,$\!^{\scriptstyle \downarrow}$ & Acc\,$\!^{\scriptstyle \uparrow}$ & Forg\,$\!^{\scriptstyle \downarrow}$ & Acc\,$\!^{\scriptstyle \uparrow}$ & Forg\,$\!^{\scriptstyle \downarrow}$ \\
    \midrule
    Full & $98.84$ & $3.96$ & $92.61$ & $6.15$ & $78.80$ & $9.89$ & $91.61$ & $11.71$ & $93.56$ & $14.18$ & $89.76$ & $4.98$ \\
    LoRA & $99.01$ & $4.34$ & $92.70$ & $6.51$ & $76.87$ & $19.71$ & $92.07$ & $15.62$ & $93.94$ & $11.58$ & $89.98$ & $6.51$ \\
    DoRA & $99.01$ & $4.15$ & $92.72$ & $6.48$ & $76.93$ & $19.88$ & $\bm{92.14}$ & $15.64$ & $93.94$ & $11.63$ & $89.99$ & $6.67$ \\
    OLoRA* & $98.71$ & $46.15$ & $92.01$ & $38.15$ & $75.88$ & $48.98$ & $91.61$ & $42.87$ & $92.71$ & $25.27$ & $89.14$ & $29.59$ \\
    \midrule
    0 (PiSSA) & $\bm{99.02}$ & $5.45$ & $93.04$ & $8.23$ & $77.46$ & $11.03$ & $91.74$ & $7.24$ & $93.87$ & $8.46$ & $89.75$ & $7.01$ \\
    \rowcolor{yellow!20}128 (16\%) & $\bm{99.02}$ & $\bm{1.48}$ & $\bm{93.11}$ & $2.98$ & $78.50$ & $4.48$ & $91.46$ & $\bm{3.36}$ & $\bm{94.56}$ & $4.53$ & $89.85$ & $2.61$ \\
    \rowcolor{yellow!20}256 (33\%) & $99.01$ & $1.62$ & $\bm{93.11}$ & $2.59$ & $78.47$ & $\bm{4.05}$ & $91.65$ & $3.99$ & $94.45$ & $\bm{4.23}$ & $89.98$ & $\bm{2.55}$ \\
    \rowcolor{yellow!20}512 (66\%) & $\bm{99.02}$ & $1.71$ & $93.08$ & $\bm{2.54}$ & $78.72$ & $5.63$ & $91.57$ & $3.48$ & $94.52$ & $4.78$ & $90.05$ & $2.75$ \\
    736 (MiLoRA) & $99.00$ & $2.14$ & $92.95$ & $3.39$ & $78.40$ & $6.20$ & $91.83$ & $5.81$ & $94.47$ & $6.43$ & $\bm{90.07}$ & $3.40$ \\
    \bottomrule
    \end{tabular}
    
    \begin{tabular}{ccccccccccccc}
    \toprule
    Methods & \multicolumn{2}{c}{O. Pets} & \multicolumn{2}{c}{O. Flowers} & \multicolumn{2}{c}{S. Cars} & \multicolumn{2}{c}{S. Dogs} & \multicolumn{2}{c}{FGVC Aircraft} & \multicolumn{2}{c}{Average} \\
     & Acc\,$\!^{\scriptstyle \uparrow}$ & Forg\,$\!^{\scriptstyle \downarrow}$ & Acc\,$\!^{\scriptstyle \uparrow}$ & Forg\,$\!^{\scriptstyle \downarrow}$ & Acc\,$\!^{\scriptstyle \uparrow}$ & Forg\,$\!^{\scriptstyle \downarrow}$ & Acc\,$\!^{\scriptstyle \uparrow}$ & Forg\,$\!^{\scriptstyle \downarrow}$ & Acc\,$\!^{\scriptstyle \uparrow}$ & Forg\,$\!^{\scriptstyle \downarrow}$ & Acc\,$\!^{\scriptstyle \uparrow}$ & Forg\,$\!^{\scriptstyle \downarrow}$ \\
    \midrule
    Full & $92.93$ & $5.30$ & $98.38$ & $3.49$ & $\bm{83.51}$ & $8.82$ & $87.58$ & $2.69$ & $\bm{73.93}$ & $6.34$ & $89.23$ & $7.05$ \\
    LoRA & $92.95$ & $6.62$ & $98.76$ & $4.49$ & $80.77$ & $12.13$ & $90.08$ & $0.43$ & $66.89$ & $8.70$ & $88.55$ & $8.79$ \\
    DoRA & $92.98$ & $6.56$ & $98.77$ & $4.52$ & $80.87$ & $12.28$ & $90.06$ & $0.44$ & $67.29$ & $8.73$ & $88.61$ & $8.82$ \\
    OLoRA* & $91.67$ & $26.81$ & $97.80$ & $28.06$ & $81.32$ & $46.05$ & $87.22$ & $10.58$ & $70.25$ & $50.76$ & $88.03$ & $35.75$ \\
    \midrule
    0 (PiSSA) & $93.28$ & $4.55$ & $99.00$ & $2.25$ & $81.83$ & $9.55$ & $89.93$ & $0.81$ & $71.00$ & $6.41$ & $89.08$ & $6.45$ \\
    \rowcolor{yellow!20}128 (16\%) & $93.65$ & $2.41$ & $99.15$ & $\bm{0.53}$ & $81.91$ & $4.11$ & $90.43$ & $0.18$ & $69.18$ & $2.94$ & $89.17$ & $2.69$ \\
    \rowcolor{yellow!20}256 (33\%) & $\bm{93.74}$ & $2.15$ & $99.13$ & $0.58$ & $82.01$ & $\bm{3.85}$ & $90.54$ & $0.06$ & $69.24$ & $\bm{2.57}$ & $89.21$ & $\bm{2.57}$ \\
    \rowcolor{yellow!20}512 (66\%) & $93.72$ & $\bm{2.09}$ & $\bm{99.17}$ & $0.71$ & $81.97$ & $4.38$ & $\bm{90.73}$ & $-0.02$ & $69.68$ & $2.92$ & $\bm{89.29}$ & $2.82$ \\
    736 (MiLoRA) & $93.39$ & $2.11$ & $99.09$ & $0.87$ & $81.14$ & $4.87$ & $90.67$ & $\bm{-0.09}$ & $68.50$ & $3.35$ & $89.05$ & $3.50$ \\
    \bottomrule
    \end{tabular}
    \label{tab:cls}
    \\ {\footnotesize 
        *Note that for all PEFT methods we use the huggingface peft library, with rank 32 and learning rate 2e-5. Results in Appendix
        
        \nointerlineskip 
        
        with higher (3e-4) and lower (5e-6) learning rates confirm the high forgetting for OLoRA.
       }
\end{table*}

\begin{table*}[t]
    \centering
    \scriptsize
    \caption{Mathematical reasoning, Python coding, and  common sense results with LLaMA-2 7b. We report mean and standard deviation over 4 independent runs. High standard deviations include runs with exploding gradients. We highlight $\textbf{best}$, \underline{second best} and in \textcolor{yellow}{yellow} the best rows overall.}
    \begin{tabular}{cccccccc}
    \toprule
    Methods & Setup & \multicolumn{2}{c}{Mathematical reasoning} & \multicolumn{2}{c}{Python coding} & \multicolumn{2}{c}{Common sense} \\
    && Acc\,$\!^{\scriptstyle \uparrow}$ & Forg\,$\!^{\scriptstyle \downarrow}$ & Acc\,$\!^{\scriptstyle \uparrow}$ & Forg\,$\!^{\scriptstyle \downarrow}$ & Acc\,$\!^{\scriptstyle \uparrow}$ & Forg\,$\!^{\scriptstyle \downarrow}$ \\
    \midrule
    DoRA & ext & $39.57_{\pm0.48}$ & $\underline{3.29}_{\pm0.01}$ & $35.46_{\pm2.07}$ & $\bm{3.13}_{\pm0.00}$ & $83.28_{\pm0.21}$ & $\bm{3.03}_{\pm0.00}$ \\
    OLoRA & ext & $38.05_{\pm0.53}$ & $4.31_{\pm0.03}$ & $32.59_{\pm1.09}$ & $3.61_{\pm0.02}$ & $83.15_{\pm0.21}$ & $3.83_{\pm0.08}$ \\
    0 (PiSSA) & pissa & $39.46_{\pm0.50}$ & $\bm{3.26}_{\pm0.00}$ & $34.31_{\pm0.56}$ & $\underline{3.19}_{\pm0.00}$ & $84.19_{\pm0.27}$ & $3.24_{\pm0.04}$ \\
    0 (PiSSA) & ext & $21.27_{\pm22.16}$ & $7.23_{\pm3.03}$ & $37.85_{\pm1.35}$ & $4.71_{\pm0.64}$ & $82.41_{\pm0.13}$ & $3.64_{\pm0.20}$ \\
    \rowcolor{yellow!20}1024 (25\%) & ext & $42.77_{\pm0.67}$ & $3.84_{\pm0.16}$ & $\underline{41.70}_{\pm1.17}$ & $3.63_{\pm0.12}$ & $\bm{84.62}_{\pm0.17}$ & $3.17_{\pm0.02}$ \\
    \rowcolor{yellow!20}2048 (50\%) & ext & $\underline{42.49}_{\pm0.69}$ & $3.64_{\pm0.09}$ & $\bm{42.41}_{\pm1.65}$ & $3.70_{\pm0.06}$ & $\underline{84.49}_{\pm0.19}$ & $\underline{3.14}_{\pm0.01}$ \\
    \rowcolor{yellow!20}3072 (75\%) & ext & $\bm{42.97}_{\pm0.25}$ & $3.61_{\pm0.07}$ & $41.18_{\pm1.37}$ & $3.44_{\pm0.01}$ & $84.38_{\pm0.20}$ & $\underline{3.14}_{\pm0.03}$ \\
    3968 (MiLoRA) & ext & $32.03_{\pm21.36}$ & $5.00_{\pm2.73}$ & $29.71_{\pm19.83}$ & $5.25_{\pm3.25}$ & $84.30_{\pm0.25}$ & $3.21_{\pm0.01}$ \\
    3968 (MiLoRA) & milora & $40.18_{\pm0.47}$ & $4.27_{\pm0.20}$ & $39.42_{\pm1.40}$ & $3.59_{\pm0.10}$ & $74.63_{\pm2.17}$ & $10.68_{\pm2.14}$ \\
    \bottomrule
    \end{tabular}
    \label{tab:llm}
\end{table*}

\subsubsection{Other model backbone}
We run our main classification experiments using the bigger DINOv2 (ViT-L) as backbone. We report the results for a variety of starting points in Table~\ref{tab:cls_dinov2}. Here the linear layer size is 1024, and show that they form the observed U-shape phenomenon, and that the best performance-trade off is achieved with intermediate components.

\subsubsection{Other ranks}
We run our main classification experiments with lower (r=8) and bigger (r=64) ranks, and report the results in the Table~\ref{tab:cls_ranks}. Results for these additional ranks confirm the observed U-shape phenomenon and that intermediate components show improved performance-forgetting trade-offs.

\subsection{Natural Language Processing Tasks}
Here, we study whether our findings generalize to NLP tasks. We fine-tune a pre-trained LLaMA-2 model on three NLP tasks: mathematical reasoning, python coding and common sense. We study the impact of fine-tuning components at the extremes (PiSSA and MiLoRA) and intermediate ones, using our generalization. We use different starting points $s$ (0, 1024, 2048, 3072, 3968) with rank $r=128$, where $s=0$ is PiSSA \citep{meng2024pissa} and $s=3968$ is MiLoRA \citep{wang2024milora}. We benchmark all methods with 3 different training setups: the one used in PiSSA \citep{meng2024pissa}, the one suggested by MiLoRA \citep{wang2024milora}, and ext, which adapts PiSSA's one to the highest learning rate possible. We notice that MiLoRA proposes various changes to PiSSA's setup, whereas we investigate the impact of the learning rate only. In these experiments, we compute forgetting as in \citep{wang2024milora,kalajdzievski2024scaling}, with a soft cross-entropy loss which uses tokens predicted by the model before fine-tuning as targets and tokens predicted by the model after fine-tuning as predictions. Training details are reported in Appendix~\ref{app:llm_details}.

Table~\ref{tab:llm} displays the summary of results for the next token prediction tasks: mathematical reasoning, python coding and common sense datasets. Here, we see that we improve the performance-forgetting trade-off over both PISSA and MiLoRA by fine-tuning intermediate components and increasing the learning rate to ``the highest stable possible". This is possible as intermediate components interfere less with the main components, as shown in Sec.~\ref{sec:analysis}. As a result, intermediate components are also more robust to higher learning rate settings than extreme components, consistently leading to better results. Tables~\ref{tab:llm_python}, \ref{tab:llm_metamath} and \ref{tab:llm_commonsense} report the detailed results.

In Figure~\ref{fig:llm} we report average accuracy and forgetting results for fine-tuning LLaMA-2 on mathematical reasoning (top row), python coding (middle row) and common sense (bottom row) using the described training setups (PiSSA, Ext, and MiLoRA). These figures show that PiSSA excels in PiSSA's setup, MiLoRA is better than PiSSA in MiLoRA's setup, but intermediate components can be even better. Most importantly, by modifying PiSSA's setup to a higher learning rate (Ext), we observe that forgetting forms the previously observed U-shape, and accuracy the opposite, i.e., a reversed U-shape, confirming intermediate components as best performance-forgetting trade-offs. Additionally, when using a high learning rate, we see that intermediate components are more robust, whereas extremes can more easily lead to exploding gradients, and consequently high damage to the prior knowledge in the original model. Lastly, our setup leads to superior results in terms of accuracy, when compared to the other setups.

\section{Conclusion}

Low-Rank Adaptation (LoRA) has become key for adapting large pre-trained models to downstream tasks. However, they face a fundamental challenge: achieving strong task-specific performance while avoiding catastrophic forgetting of pre-trained knowledge. Existing approaches offer inconsistent guidance on how to make this trade-off.

In this paper, we offer the first principled study into principle component based initialization methods for low-rank adaptation. We offer a new analysis approach that allows us to study the impact of components used for fine-tuning and of training duration. We propose to leverage intermediate components as a means of achieving superior trade-offs in learning and forgetting.

We empirically demonstrate when a model is fine-tuned long enough, its accuracy plateaus around a maximum value for any rank, whereas the forgetting forms a U-shape, where models fine-tuned using intermediate components show the least forgetting. This suggests that components at the extremes are more prone to forget than intermediate ones. 

We therefore propose to make use of intermediate components for better trading off accuracy and forgetting. Our findings pave the way for designing targeted interventions—such as selective rank pruning or direction-constrained updates—that mitigate catastrophic forgetting while preserving downstream performance, ultimately advancing the reliability of large-scale models.

\section*{Acknowledgments}
The authors gratefully acknowledge the Gauss Centre for Supercomputing e.V. (\url{www.gauss-centre.eu}) for funding this project by providing computing time through the John von Neumann Institute for Computing (NIC) on the GCS Supercomputer JUWELS \cite{kesselheim2021juwels} at J\"ulich Supercomputing Centre (JSC).


\bibliography{main}
\bibliographystyle{icml2026}

\newpage
\appendix
\onecolumn
\setcounter{table}{0}
\renewcommand{\thetable}{S\arabic{table}}
\begin{center}
\textbf{\Large Supplementary Material}
\end{center}

\section{Hardware details}
We run our classification experiments using 1 A100 with 40GB dedicated RAM. For NLP experiments we use 8 A100s with 40GB each. All experiments are run for up to 24h.

\section{Training details}

\subsection{Image Classification}
\label{app:cls_details}
We conduct all image classification experiments using a simple training procedure for ViT-B: rank $r=32$, scaling factor $\alpha=32$, AdamW optimizer (LR=$2\times10^{-5}$, weight decay=$0.01$), batch size 10, over 200 epochs. As starting components we use 0 (PiSSA), 32, 64, 128, 256, 512, 736 (MiLoRA).


\subsection{NLP Tasks}
\label{app:llm_details}
For the NLP tasks we use 3 different training setups: PiSSA \cite{meng2024pissa}, MiLoRA \cite{wang2024milora}, and ext (i.e., PiSSA with higher learning rate). Configurations are reported in Table~\ref{tab:hyperparams}. As starting components we use: 0 (PiSSA), 1024, 2048, 3072, 3968 (MiLoRA).

\begin{table}[h]
\centering
\caption{Hyperparameter configuration on the common-sense reasoning (ComR), math reasoning (MathR) and instruction-following (InsF) tasks.}
\label{tab:hyperparams}
\small
\begin{tabular}{lccc}
\toprule
\textbf{Hyperparameters} & PiSSA & Ext & MiLoRA \\
\midrule
Rank $r$ & 128 & 128 & 64 \\
$\alpha$ of PiSSA/MiLoRA & 128 & 128 & 64 \\
Dropout & -- & -- & 0.05 \\
Optimizer & AdamW & AdamW & AdamW \\
LR & Cosine & Cosine & Linear \\
LR Scheduler & 2e-5 & 3.5e-4, 3e-4, 1e-4 & 3e-4 \\
Batch size & 128 & 128 & 16 \\
Warmup ratio & 0.03 & 0.03 & - \\
Warmup steps & - & - & 100 \\
Epochs & 3 & 3 & 3 \\
Placement & query, key, value, & query, key, value, & query, key, value, \\
& output, gate, & output, gate, & MLP up, MLP down \\
& MLP up, MLP down & MLP up, MLP down & \\
\bottomrule
\end{tabular}
\end{table}

\newpage
\clearpage
\section{Additional experiments}

\subsection{Image Classification}

\begin{figure}[h]
    \centering
    \includegraphics[width=0.6\textwidth, trim=0 11cm 0 0, clip]{figures/legend_forg_all_last_new_new.pdf} 
    
    \begin{subfigure}{0.35\textwidth}
        \centering
        \includegraphics[width=\textwidth]{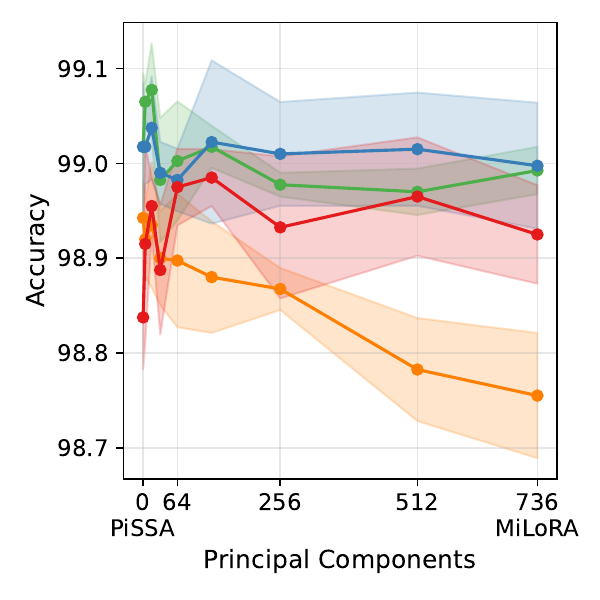}
    \end{subfigure}
    \hspace{1cm}
    \begin{subfigure}{0.35\textwidth}
        \centering
        \includegraphics[width=\textwidth]{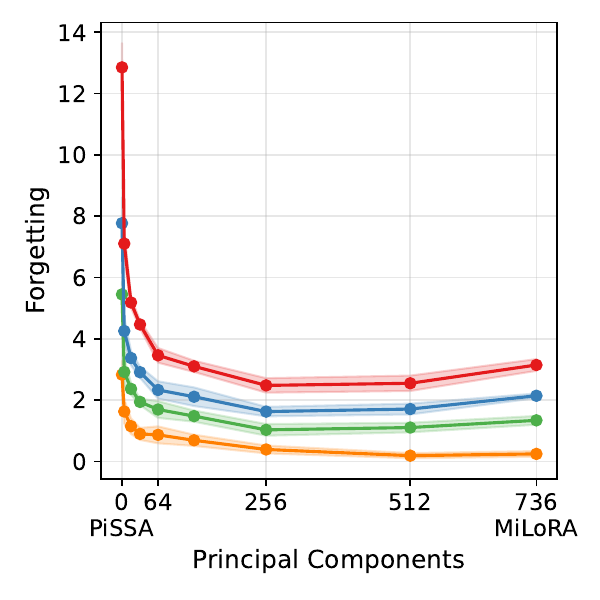}
    \end{subfigure}

    \caption{(ImageNet1k $\rightarrow$ CIFAR10) Results of fine-tuning an ImageNet1k pre-trained ViT-Base to CIFAR10 using intermediate components with rank 32, using different starting points. From left to right, accuracy of CIFAR10, and forgetting of ImageNet1k at the end of fine-tuning.}
    \label{fig:in_cifar10}
\end{figure}

\begin{figure}[h]
    \centering
    \includegraphics[width=0.6\textwidth, trim=0 11cm 0 0, clip]{figures/legend_forg_all_last_new_new.pdf} 
    
    \begin{subfigure}{0.35\textwidth}
        \centering
        \includegraphics[width=\textwidth]{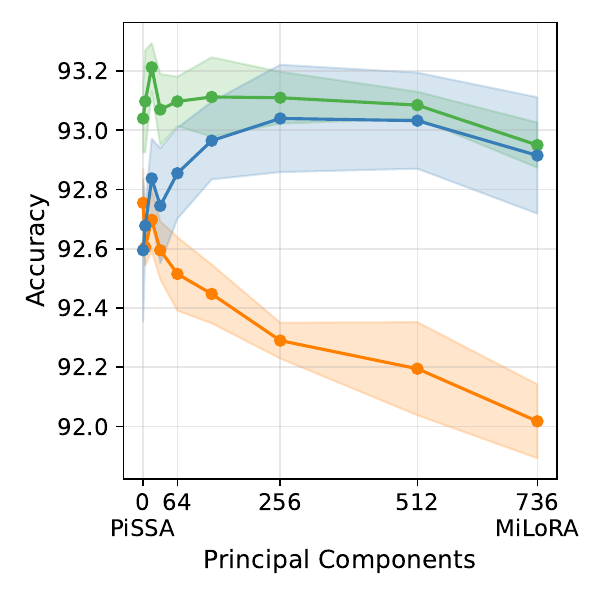}
    \end{subfigure}
    \hspace{1cm}
    \begin{subfigure}{0.35\textwidth}
        \centering
        \includegraphics[width=\textwidth]{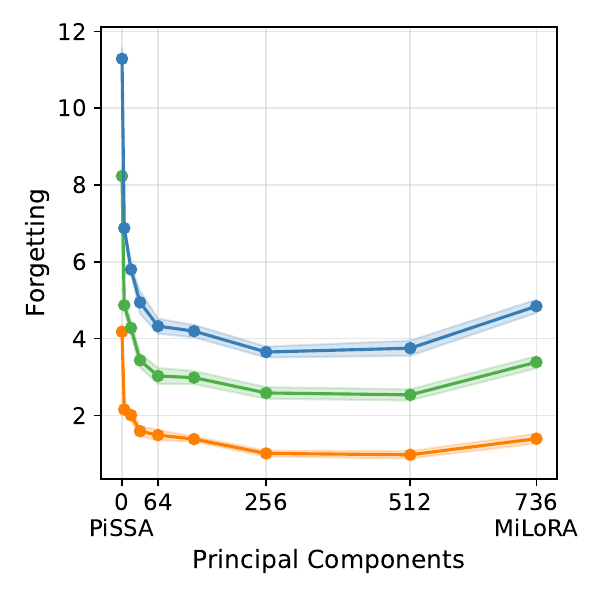}
    \end{subfigure}
    
    \caption{(ImageNet1k $\rightarrow$ CIFAR100) Results of fine-tuning an ImageNet1k pre-trained ViT-Base to CIFAR100 using intermediate components with rank 32, using different starting points. From left to right, accuracy of CIFAR100, and forgetting of ImageNet1k at the end of fine-tuning.}
    \label{fig:in_cifar100}
\end{figure}

\begin{figure}[h]
    \centering
    \includegraphics[width=0.6\textwidth, trim=0 11cm 0 0, clip]{figures/legend_forg_all_last_new_new.pdf} 
    
    \begin{subfigure}{0.35\textwidth}
        \centering
        \includegraphics[width=\textwidth]{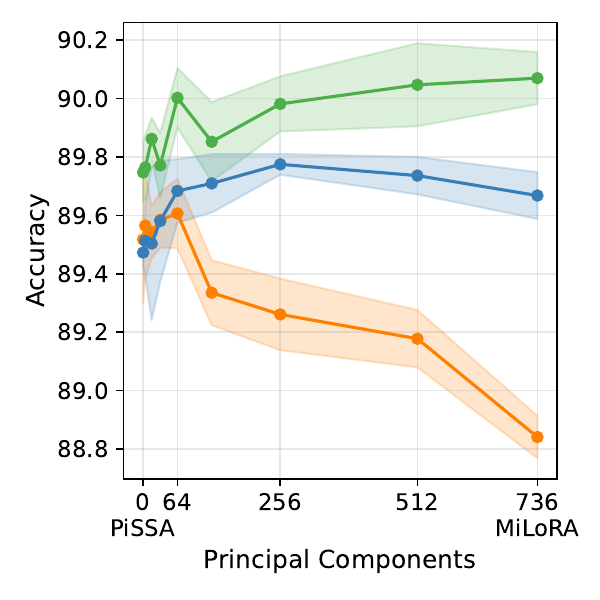}
    \end{subfigure}
    \hspace{1cm}
    \begin{subfigure}{0.35\textwidth}
        \centering
        \includegraphics[width=\textwidth]{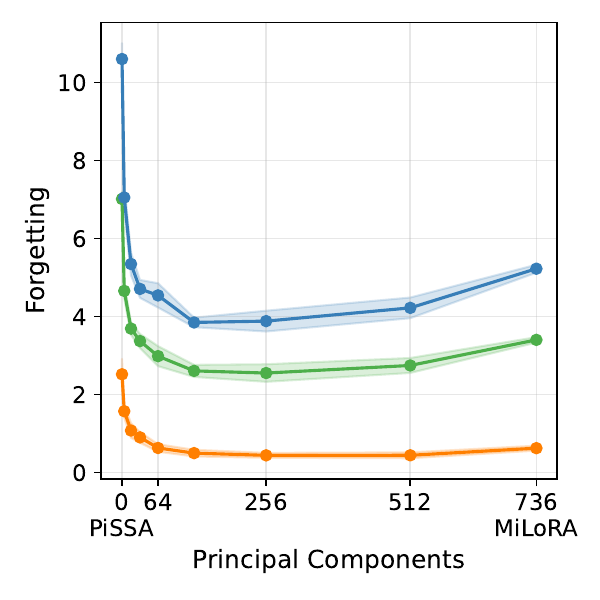}
    \end{subfigure}
    
    \caption{(ImageNet1k $\rightarrow$ Food101) Results of fine-tuning an ImageNet1k pre-trained ViT-Base to Food101 using intermediate components with rank 32, using different starting points. From left to right, accuracy of Food101, and forgetting of ImageNet1k at the end of fine-tuning.}
    \label{fig:in_food101}
\end{figure}

\begin{figure}[h]
    \centering
    \includegraphics[width=0.6\textwidth, trim=0 11cm 0 0, clip]{figures/legend_forg_all_last_new_new.pdf} 
    
    \begin{subfigure}{0.35\textwidth}
        \centering
        \includegraphics[width=\textwidth]{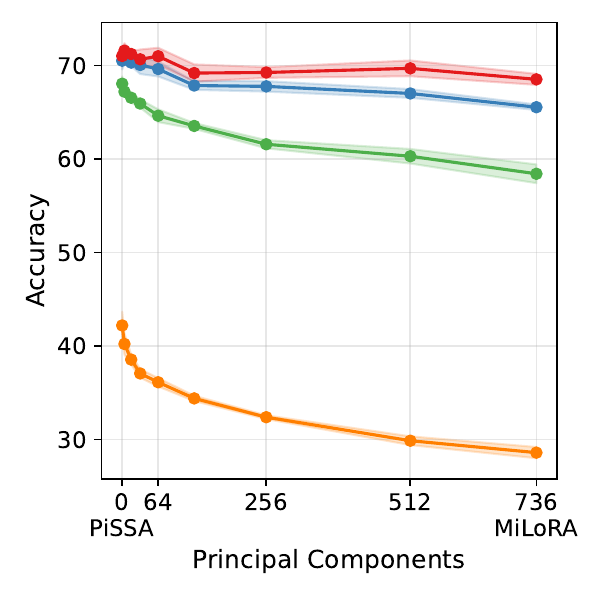}
    \end{subfigure}
    \hspace{1cm}
    \begin{subfigure}{0.35\textwidth}
        \centering
        \includegraphics[width=\textwidth]{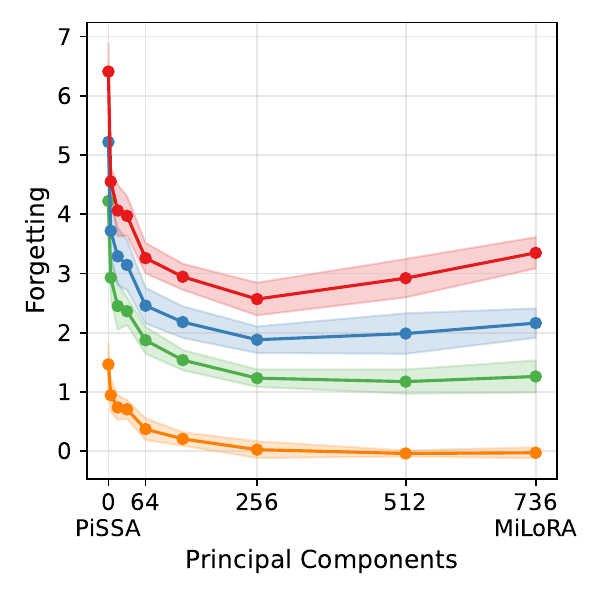}
    \end{subfigure}
    
    \caption{(ImageNet1k $\rightarrow$ FGVC Aircraft) Results of fine-tuning an ImageNet1k pre-trained ViT-Base to FGVC Aircraft using intermediate components with rank 32, using different starting points. From left to right, accuracy of FGVC Aircraft, and forgetting of ImageNet1k at the end of fine-tuning.}
    \label{fig:in_aircraft}
\end{figure}

    
    

\begin{figure}[h]
    \centering
    \includegraphics[width=0.6\textwidth, trim=0 11cm 0 0, clip]{figures/legend_forg_all_last_new_new.pdf} 
    
    \begin{subfigure}{0.35\textwidth}
        \centering
        \includegraphics[width=\textwidth]{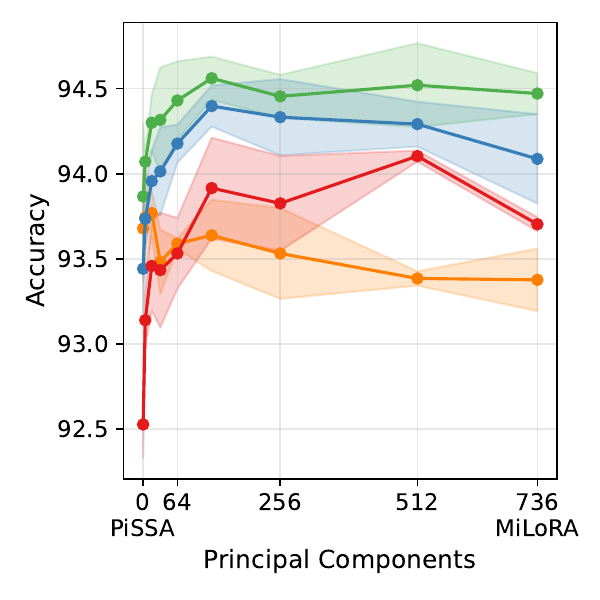}
    \end{subfigure}
    \hspace{1cm}
    \begin{subfigure}{0.35\textwidth}
        \centering
        \includegraphics[width=\textwidth]{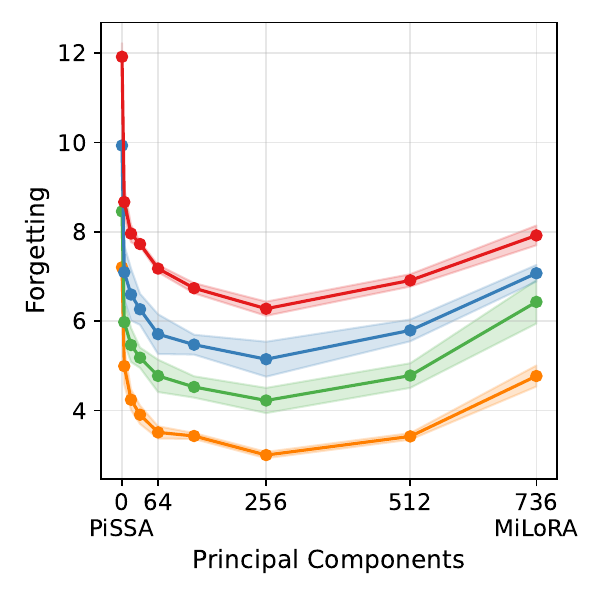}
    \end{subfigure}
    
    \caption{(ImageNet1k $\rightarrow$ Caltech256) Results of fine-tuning an ImageNet1k pre-trained ViT-Base to Caltech256 using intermediate components with rank 32, using different starting points. From left to right, accuracy of Caltech256, and forgetting of ImageNet1k at the end of fine-tuning.}
    \label{fig:in_caltech256}
\end{figure}

\begin{figure}[h]
    \centering
    \includegraphics[width=0.6\textwidth, trim=0 11cm 0 0, clip]{figures/legend_forg_all_last_new_new.pdf} 
    
    \begin{subfigure}{0.35\textwidth}
        \centering
        \includegraphics[width=\textwidth]{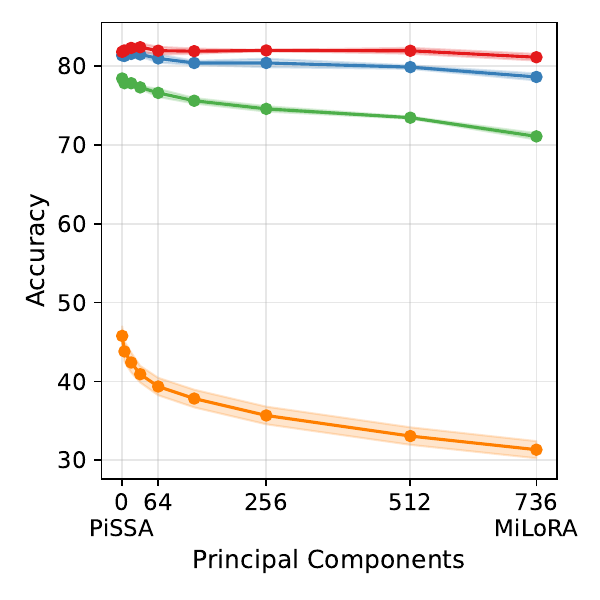}
    \end{subfigure}
    \hspace{1cm}
    \begin{subfigure}{0.35\textwidth}
        \centering
        \includegraphics[width=\textwidth]{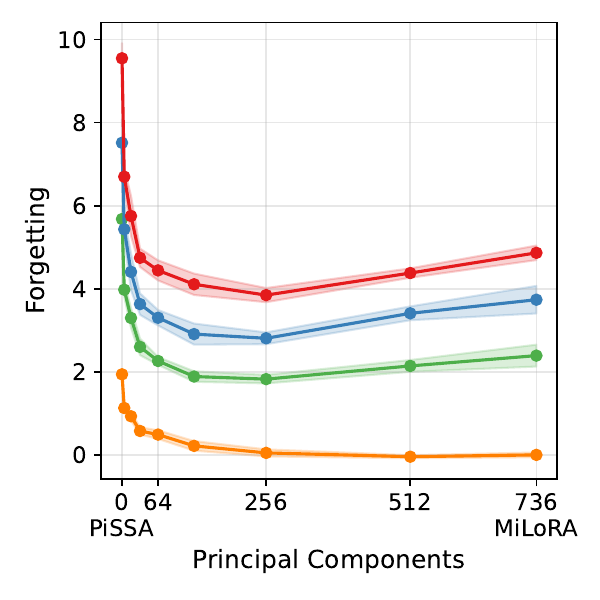}
    \end{subfigure}
    
    \caption{(ImageNet1k $\rightarrow$ Stanford-Cars) Results of fine-tuning an ImageNet1k pre-trained ViT-Base to Stanford-Cars using intermediate components with rank 32, using different starting points. From left to right, accuracy of Stanford-Cars, and forgetting of ImageNet1k at the end of fine-tuning.}
    \label{fig:in_scars}
\end{figure}

\begin{figure}[h]
    \centering
    \includegraphics[width=0.6\textwidth, trim=0 11cm 0 0, clip]{figures/legend_forg_all_last_new_new.pdf} 
    
    \begin{subfigure}{0.35\textwidth}
        \centering
        \includegraphics[width=\textwidth]{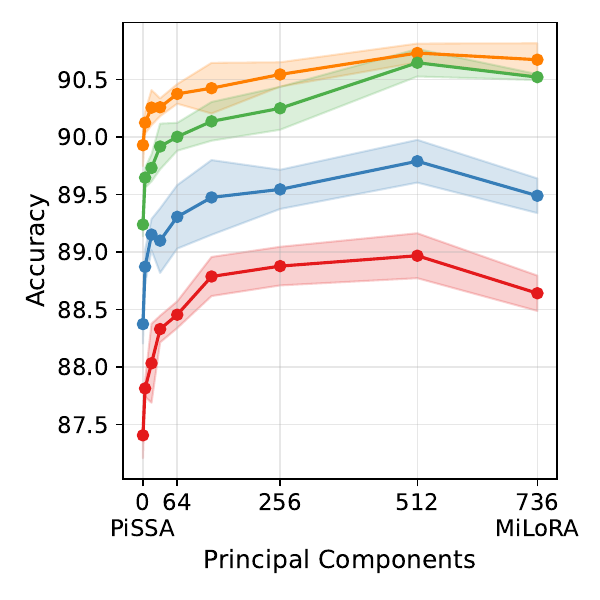}
    \end{subfigure}
    \hspace{1cm}
    \begin{subfigure}{0.35\textwidth}
        \centering
        \includegraphics[width=\textwidth]{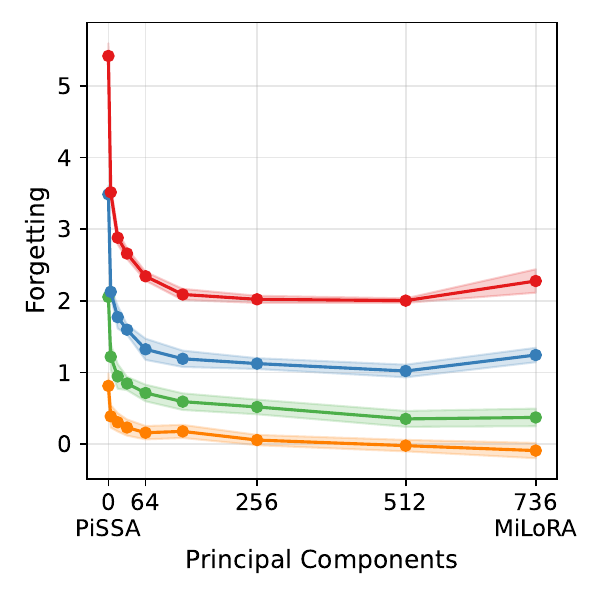}
    \end{subfigure}
    
    \caption{(ImageNet1k $\rightarrow$ Stanford-Dogs) Results of fine-tuning an ImageNet1k pre-trained ViT-Base to Stanford-Dogs using intermediate components with rank 32, using different starting points. From left to right, accuracy of Stanford-Dogs, and forgetting of ImageNet1k at the end of fine-tuning.}
    \label{fig:in_sdogs}
\end{figure}

\begin{figure}[h]
    \centering
    \includegraphics[width=0.6\textwidth, trim=0 11cm 0 0, clip]{figures/legend_forg_all_last_new_new.pdf} 
    
    \begin{subfigure}{0.35\textwidth}
        \centering
        \includegraphics[width=\textwidth]{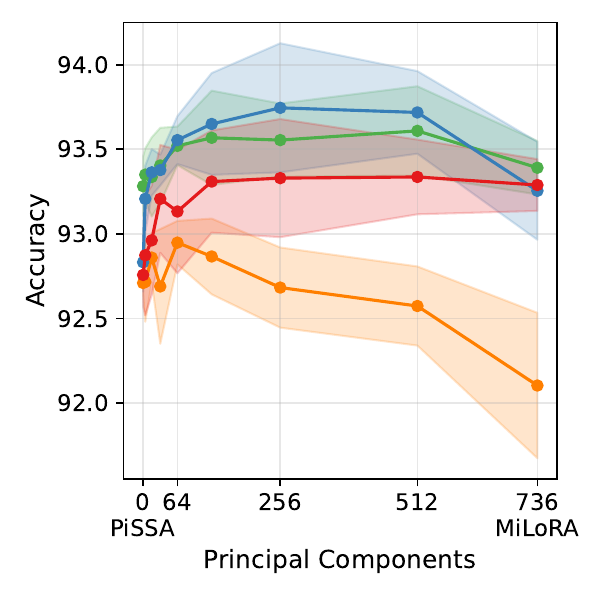}
    \end{subfigure}
    \hspace{1cm}
    \begin{subfigure}{0.35\textwidth}
        \centering
        \includegraphics[width=\textwidth]{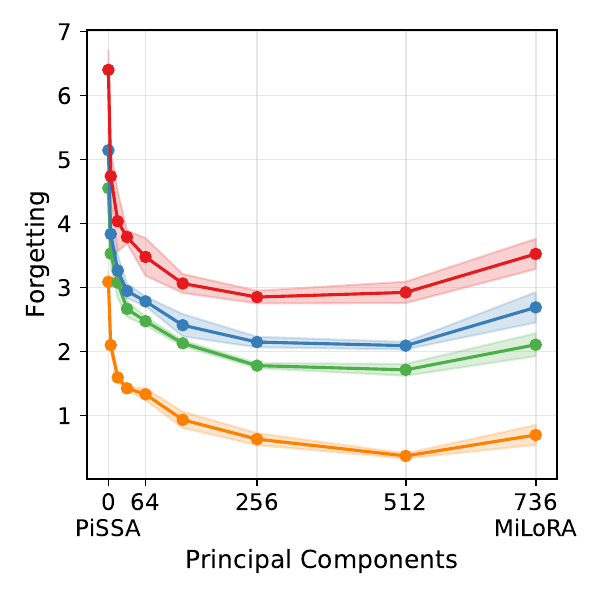}
    \end{subfigure}
    
    \caption{(ImageNet1k $\rightarrow$ Oxford Pets) Results of fine-tuning an ImageNet1k pre-trained ViT-Base to Oxford Pets using intermediate components with rank 32, using different starting points. From left to right, accuracy of Oxford Pets, and forgetting of ImageNet1k, at the end of fine-tuning.}
    \label{fig:in_o3tpet}
\end{figure}

\begin{figure}[h]
    \centering
    \includegraphics[width=0.6\textwidth, trim=0 11cm 0 0, clip]{figures/legend_forg_all_last_new_new.pdf} 
    
    \begin{subfigure}{0.35\textwidth}
        \centering
        \includegraphics[width=\textwidth]{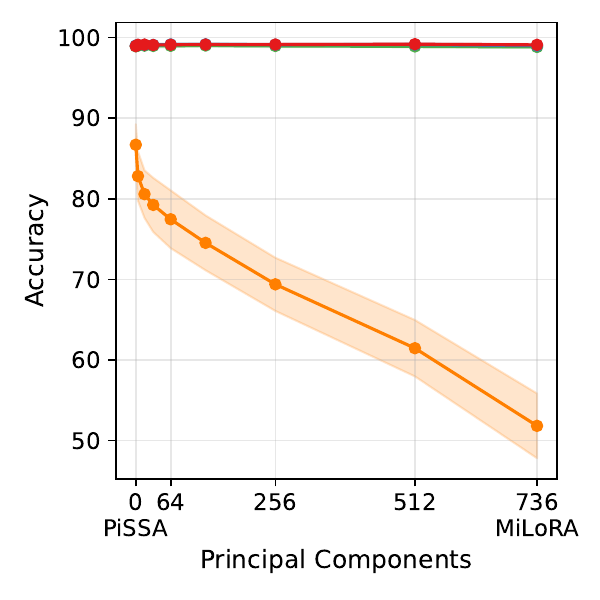}
    \end{subfigure}
    \hspace{1cm}
    \begin{subfigure}{0.35\textwidth}
        \centering
        \includegraphics[width=\textwidth]{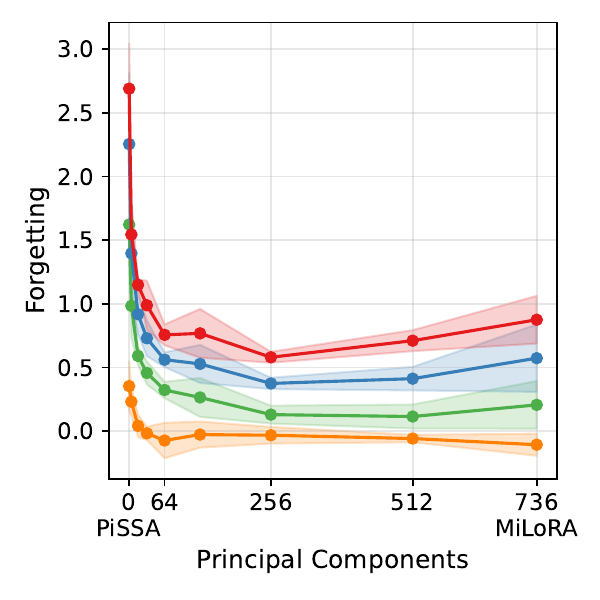}
    \end{subfigure}
    
    \caption{(ImageNet1k $\rightarrow$ Oxford Flowers102) Results of fine-tuning an ImageNet1k pre-trained ViT-Base to Oxford Flowers102 using intermediate components with rank 32, using different starting points. From left to right, accuracy of Oxford Flowers102, and forgetting of ImageNet1k at the end of fine-tuning.}
    \label{fig:in_oflowers102}
\end{figure}

\begin{figure}[h]
    \centering
    \includegraphics[width=0.6\textwidth, trim=0 11cm 0 0, clip]{figures/legend_forg_all_last_new_new.pdf} 
    
    \begin{subfigure}{0.35\textwidth}
        \centering
        \includegraphics[width=\textwidth]{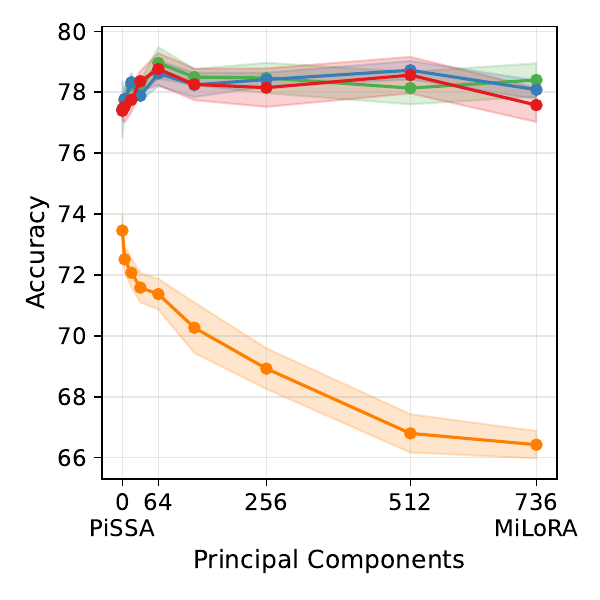}
    \end{subfigure}
    \hspace{1cm}
    \begin{subfigure}{0.35\textwidth}
        \centering
        \includegraphics[width=\textwidth]{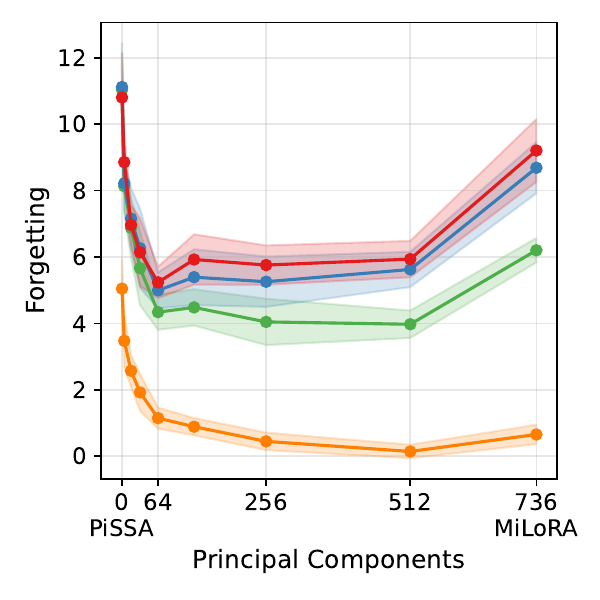}
    \end{subfigure}
    
    \caption{(ImageNet1k $\rightarrow$ DTD) Results of fine-tuning an ImageNet1k pre-trained ViT-Base to DTD using intermediate components with rank 32, using different starting points. From left to right, accuracy of DTD, and forgetting of ImageNet1k at the end of fine-tuning.}
    \label{fig:in_dtd}
\end{figure}

\begin{table*}[h]
    \setlength{\tabcolsep}{3pt} 
    \scriptsize
    \centering
    \caption{Accuracy and Forgetting of Imagenet1k and various finetuned classification datasets, after fine-tuning for up to 200 epochs. Maximum mean and standard deviation of 4 independent runs. We highlight $\textbf{best}$, and in \textcolor{yellow}{yellow} the best rows overall.}
    \begin{tabular}{ccccccc}
    \toprule
    Methods & \multicolumn{2}{c}{CIFAR10} & \multicolumn{2}{c}{CIFAR100} & \multicolumn{2}{c}{DTD} \\
     & Acc\,$\!^{\scriptstyle \uparrow}$ & Forg\,$\!^{\scriptstyle \downarrow}$ & Acc\,$\!^{\scriptstyle \uparrow}$ & Forg\,$\!^{\scriptstyle \downarrow}$ & Acc\,$\!^{\scriptstyle \uparrow}$ & Forg\,$\!^{\scriptstyle \downarrow}$ \\
    \midrule
    Full & $98.84_{\pm0.14}$ & $3.96_{\pm0.54}$ & $92.61_{\pm0.12}$ & $6.15_{\pm0.42}$ & $78.80_{\pm0.61}$ & $9.89_{\pm1.95}$ \\
    LoRA & $99.01_{\pm0.08}$ & $4.34_{\pm0.26}$ & $92.70_{\pm0.27}$ & $6.51_{\pm0.40}$ & $76.87_{\pm1.01}$ & $19.71_{\pm1.07}$ \\
    DoRA & $99.01_{\pm0.09}$ & $4.15_{\pm0.34}$ & $92.72_{\pm0.21}$ & $6.48_{\pm0.44}$ & $76.93_{\pm0.93}$ & $19.88_{\pm1.04}$ \\
    OLoRA & $98.71_{\pm0.11}$ & $46.15_{\pm2.57}$ & $92.01_{\pm0.10}$ & $38.15_{\pm1.10}$ & $75.88_{\pm0.79}$ & $48.98_{\pm1.59}$ \\
    $\text{OLoRA}_{3e-4}$ & $96.45_{\pm0.08}$ & $76.64_{\pm0.31}$ & $85.06_{\pm0.57}$ & $60.19_{\pm0.64}$ & $70.24_{\pm2.11}$ & $62.61_{\pm3.36}$ \\
    $\text{OLoRA}_{5e-6}$ & $98.91_{\pm0.03}$ & $30.68_{\pm1.91}$ & $92.39_{\pm0.09}$ & $34.94_{\pm0.99}$ & $76.74_{\pm0.49}$ & $39.73_{\pm1.70}$ \\
    \midrule
    0 (PiSSA, 0\%) & $\bm{99.02}_{\pm0.08}$ & $5.45_{\pm0.41}$ & $93.04_{\pm0.11}$ & $8.23_{\pm0.30}$ & $77.46_{\pm0.23}$ & $11.03_{\pm1.12}$ \\
    32 (4\%) & $98.99_{\pm0.03}$ & $2.91_{\pm0.18}$ & $93.07_{\pm0.12}$ & $3.43_{\pm0.18}$ & $78.36_{\pm0.33}$ & $6.14_{\pm1.02}$ \\
    64 (8\%) & $99.00_{\pm0.06}$ & $1.69_{\pm0.27}$ & $93.10_{\pm0.08}$ & $3.03_{\pm0.21}$ & $\bm{78.96}_{\pm0.52}$ & $4.34_{\pm0.53}$ \\
    \rowcolor{yellow!20}128 (16\%) & $\bm{99.02}_{\pm0.02}$ & $\bm{1.48}_{\pm0.18}$ & $\bm{93.11}_{\pm0.13}$ & $2.98_{\pm0.17}$ & $78.50_{\pm0.27}$ & $4.48_{\pm0.55}$ \\
    \rowcolor{yellow!20}256 (33\%) & $99.01_{\pm0.05}$ & $1.62_{\pm0.16}$ & $\bm{93.11}_{\pm0.09}$ & $2.59_{\pm0.15}$ & $78.47_{\pm0.50}$ & $\bm{4.05}_{\pm0.70}$ \\
    \rowcolor{yellow!20}512 (66\%) & $\bm{99.02}_{\pm0.06}$ & $1.71_{\pm0.18}$ & $93.08_{\pm0.05}$ & $\bm{2.54}_{\pm0.15}$ & $78.72_{\pm0.31}$ & $5.63_{\pm0.53}$ \\
    736 (MiLoRA, 95\%) & $99.00_{\pm0.07}$ & $2.14_{\pm0.08}$ & $92.95_{\pm0.08}$ & $3.39_{\pm0.16}$ & $78.40_{\pm0.55}$ & $6.20_{\pm0.37}$ \\
    \bottomrule
    \end{tabular}
    
    \begin{tabular}{ccccccc}
    \toprule
    Methods & \multicolumn{2}{c}{Caltech101} & \multicolumn{2}{c}{Caltech256} & \multicolumn{2}{c}{Food101} \\
     & Acc\,$\!^{\scriptstyle \uparrow}$ & Forg\,$\!^{\scriptstyle \downarrow}$ & Acc\,$\!^{\scriptstyle \uparrow}$ & Forg\,$\!^{\scriptstyle \downarrow}$ & Acc\,$\!^{\scriptstyle \uparrow}$ & Forg\,$\!^{\scriptstyle \downarrow}$ \\
    \midrule
    Full & $91.61_{\pm0.99}$ & $11.71_{\pm0.74}$ & $93.56_{\pm0.18}$ & $14.18_{\pm1.09}$ & $89.76_{\pm0.16}$ & $4.98_{\pm0.23}$ \\
    LoRA & $92.07_{\pm0.42}$ & $15.62_{\pm0.71}$ & $93.94_{\pm0.13}$ & $11.58_{\pm0.23}$ & $89.98_{\pm0.04}$ & $6.51_{\pm0.59}$ \\
    DoRA & $\bm{92.14}_{\pm0.47}$ & $15.64_{\pm0.69}$ & $93.94_{\pm0.15}$ & $11.63_{\pm0.16}$ & $89.99_{\pm0.10}$ & $6.67_{\pm0.64}$ \\
    OLoRA & $91.61_{\pm0.86}$ & $42.87_{\pm1.04}$ & $92.71_{\pm0.19}$ & $25.27_{\pm1.03}$ & $89.14_{\pm0.19}$ & $29.59_{\pm0.59}$ \\
    $\text{OLoRA}_{3e-4}$ & $86.11_{\pm3.7}$ & $44.44_{\pm2.71}$ & $86.08_{\pm0.47}$ & $33.19_{\pm0.84}$ & $83.46_{\pm0.53}$ & $60.85_{\pm1.30}$ \\
    $\text{OLoRA}_{5e-6}$ & $91.47_{\pm0.86}$ & $40.59_{\pm1.30}$ & $92.99_{\pm0.15}$ & $27.52_{\pm1.15}$ & $89.55_{\pm0.16}$ & $28.11_{\pm1.12}$ \\
    \midrule
    0 (PiSSA, 0\%) & $91.74_{\pm1.13}$ & $7.24_{\pm0.28}$ & $93.87_{\pm0.43}$ & $8.46_{\pm0.55}$ & $89.75_{\pm0.10}$ & $7.01_{\pm0.20}$ \\
    32 (4\%) & $91.65_{\pm0.59}$ & $4.21_{\pm0.18}$ & $94.32_{\pm0.31}$ & $5.18_{\pm0.23}$ & $89.77_{\pm0.11}$ & $3.37_{\pm0.18}$ \\
    64 (8\%) & $91.80_{\pm0.83}$ & $3.75_{\pm0.22}$ & $94.43_{\pm0.23}$ & $4.78_{\pm0.36}$ & $90.00_{\pm0.10}$ & $2.98_{\pm0.26}$ \\
    \rowcolor{yellow!20}128 (16\%) & $91.46_{\pm0.70}$ & $\bm{3.36}_{\pm0.18}$ & $\bm{94.56}_{\pm0.13}$ & $4.53_{\pm0.24}$ & $89.85_{\pm0.14}$ & $2.61_{\pm0.15}$ \\
    \rowcolor{yellow!20}256 (33\%) & $91.65_{\pm0.49}$ & $3.99_{\pm0.22}$ & $94.45_{\pm0.13}$ & $\bm{4.23}_{\pm0.28}$ & $89.98_{\pm0.09}$ & $\bm{2.55}_{\pm0.23}$ \\
    \rowcolor{yellow!20}512 (66\%) & $91.57_{\pm0.67}$ & $3.48_{\pm0.16}$ & $94.52_{\pm0.25}$ & $4.78_{\pm0.28}$ & $90.05_{\pm0.14}$ & $2.75_{\pm0.19}$ \\
    736 (MiLoRA, 95\%) & $91.83_{\pm0.29}$ & $5.81_{\pm0.45}$ & $94.47_{\pm0.12}$ & $6.43_{\pm0.49}$ & $\bm{90.07}_{\pm0.09}$ & $3.40_{\pm0.07}$ \\
    \bottomrule
    \end{tabular}
    
    \begin{tabular}{ccccccc}
    \toprule
    Methods & \multicolumn{2}{c}{O. Pets} & \multicolumn{2}{c}{O. Flowers} & \multicolumn{2}{c}{S. Cars} \\
     & Acc\,$\!^{\scriptstyle \uparrow}$ & Forg\,$\!^{\scriptstyle \downarrow}$ & Acc\,$\!^{\scriptstyle \uparrow}$ & Forg\,$\!^{\scriptstyle \downarrow}$ & Acc\,$\!^{\scriptstyle \uparrow}$ & Forg\,$\!^{\scriptstyle \downarrow}$ \\
    \midrule
    Full & $92.93_{\pm0.25}$ & $5.30_{\pm0.59}$ & $98.38_{\pm0.25}$ & $3.49_{\pm0.31}$ & $\bm{83.51}_{\pm0.48}$ & $8.82_{\pm0.47}$ \\
    LoRA & $92.95_{\pm0.31}$ & $6.62_{\pm0.37}$ & $98.76_{\pm0.25}$ & $4.49_{\pm0.57}$ & $80.77_{\pm0.49}$ & $12.13_{\pm0.46}$ \\
    DoRA & $92.98_{\pm0.18}$ & $6.56_{\pm0.37}$ & $98.77_{\pm0.27}$ & $4.52_{\pm0.57}$ & $80.87_{\pm0.44}$ & $12.28_{\pm0.34}$ \\
    OLoRA & $91.67_{\pm0.35}$ & $26.81_{\pm2.10}$ & $97.80_{\pm0.42}$ & $28.06_{\pm3.69}$ & $71.26_{\pm3.48}$ & $79.57_{\pm0.73}$ \\
    $\text{OLoRA}_{3e-4}$ & $85.88_{\pm1.99}$ & $47.72_{\pm1.32}$ & $96.12_{\pm0.68}$ & $37.37_{\pm1.26}$ & $81.32_{\pm0.57}$ & $46.05_{\pm1.90}$ \\
    $\text{OLoRA}_{5e-6}$ & $92.22_{\pm0.26}$ & $25.49_{\pm1.84}$ & $97.77_{\pm0.30}$ & $25.04_{\pm1.51}$ & $77.77_{\pm0.98}$ & $43.78_{\pm1.19}$ \\
    \midrule
    0 (PiSSA, 0\%) & $93.28_{\pm0.18}$ & $4.55_{\pm0.23}$ & $99.00_{\pm0.11}$ & $2.25_{\pm0.56}$ & $81.83_{\pm0.59}$ & $9.55_{\pm0.36}$ \\
    32 (4\%) & $93.40_{\pm0.22}$ & $2.67_{\pm0.12}$ & $99.08_{\pm0.16}$ & $0.73_{\pm0.14}$ & $82.41_{\pm0.51}$ & $4.75_{\pm0.22}$ \\
    64 (8\%) & $93.55_{\pm0.14}$ & $2.78_{\pm0.08}$ & $99.14_{\pm0.12}$ & $0.56_{\pm0.06}$ & $81.97_{\pm0.49}$ & $4.45_{\pm0.24}$ \\
    \rowcolor{yellow!20}\rowcolor{yellow!20}128 (16\%) & $93.65_{\pm0.30}$ & $2.41_{\pm0.17}$ & $99.15_{\pm0.12}$ & $\bm{0.53}_{\pm0.15}$ & $81.91_{\pm0.33}$ & $4.11_{\pm0.26}$ \\
    \rowcolor{yellow!20}\rowcolor{yellow!20}256 (33\%) & $\bm{93.74}_{\pm0.38}$ & $2.15_{\pm0.08}$ & $99.13_{\pm0.10}$ & $0.58_{\pm0.04}$ & $82.01_{\pm0.07}$ & $\bm{3.85}_{\pm0.17}$ \\
    \rowcolor{yellow!20}\rowcolor{yellow!20}512 (66\%) & $93.72_{\pm0.24}$ & $\bm{2.09}_{\pm0.06}$ & $\bm{99.17}_{\pm0.08}$ & $0.71_{\pm0.08}$ & $81.97_{\pm0.37}$ & $4.38_{\pm0.11}$ \\
    736 (MiLoRA, 95\%) & $93.39_{\pm0.16}$ & $2.11_{\pm0.18}$ & $99.09_{\pm0.12}$ & $0.87_{\pm0.19}$ & $81.14_{\pm0.42}$ & $4.87_{\pm0.17}$ \\
    \bottomrule
    \end{tabular}
    
    \begin{tabular}{ccccccc}
    \toprule
    Methods & \multicolumn{2}{c}{S. Dogs} & \multicolumn{2}{c}{FGVC Aircraft} & \multicolumn{2}{c}{Average} \\
     & Acc\,$\!^{\scriptstyle \uparrow}$ & Forg\,$\!^{\scriptstyle \downarrow}$ & Acc\,$\!^{\scriptstyle \uparrow}$ & Forg\,$\!^{\scriptstyle \downarrow}$ & Acc\,$\!^{\scriptstyle \uparrow}$ & Forg\,$\!^{\scriptstyle \downarrow}$ \\
    \midrule
    Full & $87.58_{\pm0.21}$ & $2.69_{\pm0.13}$ & $\bm{73.93}_{\pm0.46}$ & $6.34_{\pm0.80}$ & $89.23_{\pm0.35}$ & $7.05_{\pm0.66}$ \\
    LoRA & $90.08_{\pm0.29}$ & $0.43_{\pm0.11}$ & $66.89_{\pm0.72}$ & $8.70_{\pm0.89}$ & $88.55_{\pm0.36}$ & $8.79_{\pm0.51}$ \\
    DoRA & $90.06_{\pm0.34}$ & $0.44_{\pm0.08}$ & $67.29_{\pm0.96}$ & $8.73_{\pm0.93}$ & $88.61_{\pm0.38}$ & $8.82_{\pm0.51}$ \\
    OLoRA & $87.22_{\pm0.25}$ & $10.58_{\pm0.57}$ & $70.25_{\pm1.22}$ & $50.76_{\pm1.10}$ & $88.03_{\pm0.46}$ & $35.75_{\pm1.57}$ \\
    $\text{OLoRA}_{3e-4}$ & $75.53_{\pm1.11}$ & $38.97_{\pm1.63}$ & $66.82_{\pm1.15}$ & $75.86_{\pm1.34}$ & $83.01_{\pm1.18}$ & $53.08_{\pm1.51}$ \\
    $\text{OLoRA}_{5e-6}$ & $88.09_{\pm0.17}$ & $9.80_{\pm0.36}$ & $66.07_{\pm1.44}$ & $42.55_{\pm1.55}$ & $87.63_{\pm0.40}$ & $31.66_{\pm1.24}$ \\
    \midrule
    0 (PiSSA, 0\%) & $89.93_{\pm0.18}$ & $0.81_{\pm0.18}$ & $71.00_{\pm0.89}$ & $6.41_{\pm0.48}$ & $89.08_{\pm0.37}$ & $6.45_{\pm0.42}$ \\
    32 (4\%) & $90.26_{\pm0.08}$ & $0.23_{\pm0.11}$ & $70.65_{\pm1.05}$ & $3.97_{\pm0.34}$ & $89.27_{\pm0.32}$ & $3.42_{\pm0.26}$ \\
    64 (8\%) & $90.38_{\pm0.09}$ & $0.16_{\pm0.09}$ & $70.98_{\pm0.90}$ & $3.26_{\pm0.26}$ & $\bm{89.39}_{\pm0.32}$ & $2.89_{\pm0.23}$ \\
    \rowcolor{yellow!20}128 (16\%) & $90.43_{\pm0.22}$ & $0.18_{\pm0.09}$ & $69.18_{\pm0.92}$ & $2.94_{\pm0.22}$ & $89.17_{\pm0.30}$ & $2.69_{\pm0.21}$ \\
    \rowcolor{yellow!20}256 (33\%) & $90.54_{\pm0.11}$ & $0.06_{\pm0.07}$ & $69.24_{\pm0.57}$ & $\bm{2.57}_{\pm0.28}$ & $89.21_{\pm0.23}$ & $\bm{2.57}_{\pm0.22}$ \\
    \rowcolor{yellow!20}512 (66\%) & $\bm{90.73}_{\pm0.08}$ & $-0.02_{\pm0.08}$ & $69.68_{\pm0.84}$ & $2.92_{\pm0.32}$ & $89.29_{\pm0.28}$ & $2.82_{\pm0.19}$ \\
    736 (MiLoRA, 95\%) & $90.67_{\pm0.14}$ & $\bm{-0.09}_{\pm0.11}$ & $68.50_{\pm0.59}$ & $3.35_{\pm0.26}$ & $89.05_{\pm0.24}$ & $3.50_{\pm0.23}$ \\
    \bottomrule
    \end{tabular}
    \label{tab:cls_all}
\end{table*}

\newpage
\clearpage
\subsubsection{Experiments on ranks}
\begin{table*}[h]
    \centering
    \scriptsize
    \caption{Accuracy and Forgetting of Imagenet1k and various finetuned classification datasets, after fine-tuning for up to 200 epochs. Maximum mean and standard deviation of 4 independent runs. Ablation with ranks r=8 and r=64.}
    \begin{tabular}{cccccccc}
    \toprule
    Methods & Rank & \multicolumn{2}{c}{CIFAR10} & \multicolumn{2}{c}{CIFAR100} & \multicolumn{2}{c}{DTD} \\
     &  & Acc\,$\!^{\scriptstyle \uparrow}$ & Forg\,$\!^{\scriptstyle \downarrow}$ & Acc\,$\!^{\scriptstyle \uparrow}$ & Forg\,$\!^{\scriptstyle \downarrow}$ & Acc\,$\!^{\scriptstyle \uparrow}$ & Forg\,$\!^{\scriptstyle \downarrow}$ \\
    \midrule
    0 (0\%, PiSSA) & 8 & $99.03_{\pm0.02}$ & $6.51_{\pm0.10}$ & $93.12_{\pm0.10}$ & $6.01_{\pm0.17}$ & $77.41_{\pm0.81}$ & $8.59_{\pm0.64}$ \\
    \rowcolor{yellow!20}32 (4\%)& 8 & $99.02_{\pm0.03}$ & $2.57_{\pm0.16}$ & $\bm{93.24}_{\pm0.01}$ & $\bm{2.39}_{\pm0.08}$ & $78.82_{\pm0.70}$ & $2.76_{\pm0.29}$ \\
    \rowcolor{yellow!20}64 (8\%) & 8 & $99.02_{\pm0.08}$ & $1.59_{\pm0.10}$ & $93.16_{\pm0.09}$ & $2.78_{\pm0.11}$ & $\bm{78.98}_{\pm0.52}$ & $3.36_{\pm0.26}$ \\
    \rowcolor{yellow!20}128 (16\%) & 8 & $98.97_{\pm0.06}$ & $1.64_{\pm0.21}$ & $93.05_{\pm0.12}$ & $2.55_{\pm0.06}$ & $78.90_{\pm0.39}$ & $3.39_{\pm0.59}$ \\
\rowcolor{yellow!20}256 (33\%) & 8 & $\bm{99.05}_{\pm0.07}$ & $\bm{1.43}_{\pm0.19}$ & $93.16_{\pm0.19}$ & $3.06_{\pm0.14}$ & $78.80_{\pm0.71}$ & $\bm{2.64}_{\pm0.46}$ \\
    512 (66\%) & 8 & $99.02_{\pm0.06}$ & $1.89_{\pm0.18}$ & $93.09_{\pm0.12}$ & $3.47_{\pm0.12}$ & $78.86_{\pm0.81}$ & $5.35_{\pm0.61}$ \\
    736 (95\%, MiLoRA) & 8 & $99.04_{\pm0.06}$ & $2.19_{\pm0.23}$ & $92.76_{\pm0.11}$ & $2.62_{\pm0.20}$ & $78.72_{\pm0.64}$ & $5.78_{\pm0.56}$ \\
    0 (0\%, PiSSA) & 64 & $98.94_{\pm0.05}$ & $6.44_{\pm0.24}$ & $92.74_{\pm0.09}$ & $5.27_{\pm0.34}$ & $78.23_{\pm0.74}$ & $9.74_{\pm0.27}$ \\
    32 (4\%)& 64 & $99.00_{\pm0.06}$ & $2.88_{\pm0.21}$ & $92.89_{\pm0.18}$ & $4.98_{\pm0.18}$ & $78.74_{\pm0.60}$ & $6.56_{\pm0.46}$ \\
    64 (8\%) & 64 & $99.04_{\pm0.04}$ & $2.31_{\pm0.17}$ & $92.94_{\pm0.11}$ & $4.32_{\pm0.39}$ & $78.72_{\pm0.59}$ & $6.27_{\pm0.49}$ \\
    \rowcolor{yellow!20}128 (16\%) & 64 & $99.01_{\pm0.03}$ & $1.88_{\pm0.05}$ & $93.05_{\pm0.21}$ & $3.75_{\pm0.35}$ & $78.71_{\pm0.51}$ & $5.79_{\pm0.17}$ \\
    \rowcolor{yellow!20}256 (33\%) & 64 & $99.03_{\pm0.02}$ & $1.46_{\pm0.03}$ & $93.18_{\pm0.10}$ & $3.11_{\pm0.31}$ & $79.12_{\pm0.31}$ & $5.38_{\pm0.30}$ \\
    \rowcolor{yellow!20}512 (66\%) & 64 & $\bm{99.06}_{\pm0.03}$ & $\bm{1.24}_{\pm0.15}$ & $\bm{93.19}_{\pm0.15}$ & $\bm{2.90}_{\pm0.30}$ & $\bm{79.15}_{\pm0.39}$ & $\bm{5.01}_{\pm0.15}$ \\
    736 (95\%, MiLoRA) & 64 & $98.99_{\pm0.06}$ & $2.12_{\pm0.14}$ & $93.03_{\pm0.04}$ & $3.46_{\pm0.32}$ & $78.52_{\pm0.42}$ & $7.80_{\pm0.43}$ \\
    \bottomrule
    \end{tabular}
    
    \begin{tabular}{cccccccc}
    \toprule
    Methods & Rank & \multicolumn{2}{c}{Caltech101} & \multicolumn{2}{c}{Caltech256} & \multicolumn{2}{c}{Food101} \\
     &  & Acc\,$\!^{\scriptstyle \uparrow}$ & Forg\,$\!^{\scriptstyle \downarrow}$ & Acc\,$\!^{\scriptstyle \uparrow}$ & Forg\,$\!^{\scriptstyle \downarrow}$ & Acc\,$\!^{\scriptstyle \uparrow}$ & Forg\,$\!^{\scriptstyle \downarrow}$ \\
    \midrule
    0 (0\%, PiSSA) & 8 & $\bm{92.60}_{\pm0.39}$ & $5.55_{\pm0.76}$ & $94.38_{\pm0.10}$ & $4.54_{\pm0.26}$ & $90.03_{\pm0.07}$ & $4.94_{\pm0.29}$ \\
    \rowcolor{yellow!20}32 (4\%)& 8 & $92.46_{\pm0.27}$ & $\bm{1.57}_{\pm0.06}$ & $94.68_{\pm0.23}$ & $2.00_{\pm0.17}$ & $\bm{90.07}_{\pm0.08}$ & $0.91_{\pm0.03}$ \\
    \rowcolor{yellow!20}64 (8\%) & 8 & $91.95_{\pm0.35}$ & $\bm{1.57}_{\pm0.10}$ & $94.47_{\pm0.25}$ & $2.08_{\pm0.22}$ & $\bm{90.26}_{\pm0.13}$ & $0.93_{\pm0.12}$ \\
    \rowcolor{yellow!20}128 (16\%) & 8 & $91.79_{\pm0.28}$ & $1.84_{\pm0.11}$ & $94.50_{\pm0.16}$ & $\bm{1.56}_{\pm0.26}$ & $90.03_{\pm0.04}$ & $\bm{0.86}_{\pm0.04}$ \\
    \rowcolor{yellow!20}256 (33\%) & 8 & $91.79_{\pm0.46}$ & $2.11_{\pm0.20}$ & $\bm{94.71}_{\pm0.09}$ & $2.37_{\pm0.09}$ & $90.11_{\pm0.09}$ & $1.31_{\pm0.09}$ \\
    512 (66\%) & 8 & $92.11_{\pm0.59}$ & $3.01_{\pm0.15}$ & $94.57_{\pm0.26}$ & $3.21_{\pm0.21}$ & $90.23_{\pm0.06}$ & $1.46_{\pm0.07}$ \\
    736 (95\%, MiLoRA) & 8 & $92.20_{\pm0.67}$ & $4.77_{\pm0.39}$ & $94.43_{\pm0.19}$ & $3.79_{\pm0.27}$ & $90.02_{\pm0.06}$ & $2.29_{\pm0.14}$ \\
    0 (0\%, PiSSA) & 64 & $\bm{92.65}_{\pm0.81}$ & $10.26_{\pm0.25}$ & $93.83_{\pm0.32}$ & $9.80_{\pm0.16}$ & $89.84_{\pm0.16}$ & $3.34_{\pm0.41}$ \\
    32 (4\%)& 64 & $\bm{92.60}_{\pm0.80}$ & $6.40_{\pm0.48}$ & $93.99_{\pm0.24}$ & $8.24_{\pm0.12}$ & $\bm{90.07}_{\pm0.14}$ & $1.69_{\pm0.22}$ \\
    64 (8\%) & 64 & $92.19_{\pm0.57}$ & $5.65_{\pm0.24}$ & $94.07_{\pm0.25}$ & $7.47_{\pm0.36}$ & $\bm{90.07}_{\pm0.16}$ & $1.40_{\pm0.12}$ \\
    \rowcolor{yellow!20}128 (16\%) & 64 & $92.33_{\pm0.71}$ & $\bm{4.73}_{\pm0.19}$ & $94.12_{\pm0.10}$ & $6.99_{\pm0.22}$ & $89.95_{\pm0.07}$ & $\bm{1.12}_{\pm0.08}$ \\
    \rowcolor{yellow!20}256 (33\%) & 64 & $92.33_{\pm0.76}$ & $5.04_{\pm0.29}$ & $\bm{94.36}_{\pm0.17}$ & $6.32_{\pm0.12}$ & $89.95_{\pm0.04}$ & $3.53_{\pm0.23}$ \\
    \rowcolor{yellow!20}512 (66\%) & 64 & $92.40_{\pm0.82}$ & $4.95_{\pm0.28}$ & $94.29_{\pm0.07}$ & $\bm{6.06}_{\pm0.37}$ & $89.99_{\pm0.20}$ & $3.62_{\pm0.20}$ \\
    736 (95\%, MiLoRA) & 64 & $92.23_{\pm0.87}$ & $6.64_{\pm0.79}$ & $94.34_{\pm0.21}$ & $7.19_{\pm0.27}$ & $89.98_{\pm0.08}$ & $4.04_{\pm0.10}$ \\
    \bottomrule
    \end{tabular}
    
    \begin{tabular}{cccccccc}
    \toprule
    Methods & Rank & \multicolumn{2}{c}{O. Pets} & \multicolumn{2}{c}{O. Flowers} & \multicolumn{2}{c}{S. Cars} \\
     &  & Acc\,$\!^{\scriptstyle \uparrow}$ & Forg\,$\!^{\scriptstyle \downarrow}$ & Acc\,$\!^{\scriptstyle \uparrow}$ & Forg\,$\!^{\scriptstyle \downarrow}$ & Acc\,$\!^{\scriptstyle \uparrow}$ & Forg\,$\!^{\scriptstyle \downarrow}$ \\
    \midrule
    0 (0\%, PiSSA) & 8 & $93.40_{\pm0.42}$ & $2.99_{\pm0.29}$ & $99.12_{\pm0.06}$ & $1.15_{\pm0.24}$ & $80.44_{\pm0.44}$ & $8.54_{\pm0.37}$ \\
    \rowcolor{yellow!20}32 (4\%)& 8 & $\bm{93.86}_{\pm0.21}$ & $\bm{0.66}_{\pm0.10}$ & $99.20_{\pm0.02}$ & $0.18_{\pm0.10}$ & $80.58_{\pm0.29}$ & $2.48_{\pm0.21}$ \\
    \rowcolor{yellow!20}64 (8\%) & 8 & $93.83_{\pm0.27}$ & $1.49_{\pm0.23}$ & $99.23_{\pm0.03}$ & $\bm{0.02}_{\pm0.12}$ & $\bm{80.71}_{\pm0.28}$ & $\bm{2.29}_{\pm0.11}$ \\
    \rowcolor{yellow!20}128 (16\%) & 8 & $93.79_{\pm0.19}$ & $0.98_{\pm0.08}$ & $99.20_{\pm0.07}$ & $0.03_{\pm0.04}$ & $80.19_{\pm0.17}$ & $2.77_{\pm0.16}$ \\
    \rowcolor{yellow!20}256 (33\%) & 8 & $93.83_{\pm0.12}$ & $1.57_{\pm0.16}$ & $\bm{99.15}_{\pm0.04}$ & $0.14_{\pm0.12}$ & $80.17_{\pm0.32}$ & $3.21_{\pm0.34}$ \\
    512 (66\%) & 8 & $93.62_{\pm0.21}$ & $1.28_{\pm0.17}$ & $\bm{99.27}_{\pm0.02}$ & $0.17_{\pm0.08}$ & $80.22_{\pm0.22}$ & $3.84_{\pm0.27}$ \\
    736 (95\%, MiLoRA) & 8 & $93.57_{\pm0.20}$ & $1.28_{\pm0.09}$ & $\bm{99.15}_{\pm0.02}$ & $0.25_{\pm0.06}$ & $79.76_{\pm0.53}$ & $4.77_{\pm0.10}$ \\
    0 (0\%, PiSSA) & 64 & $92.86_{\pm0.50}$ & $5.44_{\pm0.54}$ & $98.81_{\pm0.20}$ & $2.93_{\pm0.21}$ & $82.56_{\pm0.38}$ & $8.04_{\pm0.83}$ \\
    32 (4\%)& 64 & $93.33_{\pm0.20}$ & $3.91_{\pm0.11}$ & $98.98_{\pm0.16}$ & $1.49_{\pm0.08}$ & $\bm{82.82}_{\pm0.15}$ & $5.00_{\pm0.15}$ \\
    64 (8\%) & 64 & $93.23_{\pm0.15}$ & $3.46_{\pm0.14}$ & $98.96_{\pm0.24}$ & $1.24_{\pm0.11}$ & $82.54_{\pm0.30}$ & $\bm{4.37}_{\pm0.26}$ \\
    \rowcolor{yellow!20}128 (16\%) & 64 & $93.43_{\pm0.10}$ & $3.06_{\pm0.06}$ & $99.03_{\pm0.18}$ & $1.10_{\pm0.08}$ & $82.40_{\pm0.61}$ & $5.34_{\pm0.17}$ \\
    \rowcolor{yellow!20}256 (33\%) & 64 & $\bm{93.59}_{\pm0.14}$ & $\bm{2.53}_{\pm0.11}$ & $99.09_{\pm0.12}$ & $0.92_{\pm0.14}$ & $82.43_{\pm0.52}$ & $4.57_{\pm0.26}$ \\
    \rowcolor{yellow!20}512 (66\%) & 64 & $\bm{93.59}_{\pm0.36}$ & $2.74_{\pm0.23}$ & $\bm{99.15}_{\pm0.14}$ & $\bm{0.85}_{\pm0.09}$ & $82.38_{\pm0.46}$ & $4.88_{\pm0.20}$ \\
    736 (95\%, MiLoRA) & 64 & $93.46_{\pm0.24}$ & $2.72_{\pm0.16}$ & $99.13_{\pm0.08}$ & $1.04_{\pm0.07}$ & $82.01_{\pm0.57}$ & $5.28_{\pm0.29}$ \\
    \bottomrule
    \end{tabular}
    
    \begin{tabular}{cccccccc}
    \toprule
    Methods & Rank & \multicolumn{2}{c}{S. Dogs} & \multicolumn{2}{c}{FGVC Aircraft} & \multicolumn{2}{c}{Average} \\
     &  & Acc\,$\!^{\scriptstyle \uparrow}$ & Forg\,$\!^{\scriptstyle \downarrow}$ & Acc\,$\!^{\scriptstyle \uparrow}$ & Forg\,$\!^{\scriptstyle \downarrow}$ & Acc\,$\!^{\scriptstyle \uparrow}$ & Forg\,$\!^{\scriptstyle \downarrow}$ \\
    \midrule
    0 (0\%, PiSSA) & 8 & $90.59_{\pm0.13}$ & $0.93_{\pm0.09}$ & $\bm{67.79}_{\pm1.25}$ & $5.50_{\pm0.30}$ & $88.90_{\pm0.34}$ & $5.02_{\pm0.32}$ \\
    \rowcolor{yellow!20}32 (4\%)& 8 & $91.22_{\pm0.09}$ & $-0.04_{\pm0.07}$ & $67.24_{\pm0.54}$ & $1.73_{\pm0.11}$ & $\bm{89.13}_{\pm0.22}$ & $\bm{1.56}_{\pm0.13}$ \\
    \rowcolor{yellow!20}64 (8\%) & 8 & $91.32_{\pm0.11}$ & $-0.08_{\pm0.02}$ & $66.85_{\pm0.81}$ & $1.72_{\pm0.20}$ & $89.07_{\pm0.27}$ & $1.61_{\pm0.14}$ \\
    \rowcolor{yellow!20}128 (16\%) & 8 & $91.21_{\pm0.14}$ & $-0.07_{\pm0.05}$ & $67.24_{\pm0.89}$ & $\bm{1.71}_{\pm0.17}$ & $88.99_{\pm0.23}$ & $1.57_{\pm0.16}$ \\
    \rowcolor{yellow!20}256 (33\%) & 8 & $91.19_{\pm0.12}$ & $\bm{-0.09}_{\pm0.03}$ & $66.31_{\pm0.54}$ & $1.76_{\pm0.16}$ & $88.93_{\pm0.25}$ & $1.77_{\pm0.18}$ \\
    512 (66\%) & 8 & $\bm{91.40}_{\pm0.07}$ & $-0.08_{\pm0.02}$ & $66.61_{\pm0.73}$ & $2.35_{\pm0.16}$ & $89.00_{\pm0.29}$ & $2.36_{\pm0.19}$ \\
    736 (95\%, MiLoRA) & 8 & $91.28_{\pm0.11}$ & $-0.08_{\pm0.03}$ & $66.98_{\pm0.73}$ & $2.93_{\pm0.18}$ & $88.90_{\pm0.30}$ & $2.78_{\pm0.20}$ \\
    0 (0\%, PiSSA) & 64 & $89.35_{\pm0.09}$ & $1.36_{\pm0.11}$ & $\bm{72.63}_{\pm0.29}$ & $7.22_{\pm0.95}$ & $89.31_{\pm0.33}$ & $6.35_{\pm0.39}$ \\
    32 (4\%) & 64 & $89.78_{\pm0.22}$ & $0.74_{\pm0.09}$ & $72.17_{\pm0.97}$ & $4.84_{\pm0.21}$ & $\bm{89.49}_{\pm0.34}$ & $4.25_{\pm0.21}$ \\
    64 (8\%) & 64 & $89.86_{\pm0.16}$ & $0.56_{\pm0.12}$ & $71.67_{\pm1.35}$ & $4.08_{\pm0.28}$ & $89.39_{\pm0.36}$ & $3.74_{\pm0.24}$ \\
    \rowcolor{yellow!20}128 (16\%) & 64 & $89.96_{\pm0.22}$ & $0.55_{\pm0.05}$ & $71.13_{\pm0.52}$ & $3.47_{\pm0.12}$ & $89.37_{\pm0.30}$ & $3.43_{\pm0.14}$ \\
    \rowcolor{yellow!20}256 (33\%) & 64 & $90.19_{\pm0.23}$ & $0.40_{\pm0.15}$ & $69.67_{\pm1.79}$ & $\bm{2.27}_{\pm0.08}$ & $89.36_{\pm0.38}$ & $\bm{3.23}_{\pm0.18}$ \\
    \rowcolor{yellow!20}512 (66\%) & 64 & $90.29_{\pm0.08}$ & $0.24_{\pm0.09}$ & $69.68_{\pm0.70}$ & $3.15_{\pm0.19}$ & $89.38_{\pm0.31}$ & $3.24_{\pm0.20}$ \\
    736 (95\%, MiLoRA) & 64 & $\bm{90.52}_{\pm0.18}$ & $\bm{0.13}_{\pm0.05}$ & $68.85_{\pm0.67}$ & $3.40_{\pm0.44}$ & $89.19_{\pm0.31}$ & $3.98_{\pm0.28}$ \\
    \bottomrule
    \end{tabular}
    \label{tab:cls_ranks}
\end{table*}

\newpage
\clearpage
\subsubsection{Experiments on DINOv2}


\begin{table*}[h]
    \centering
    \small
    \caption{Accuracy and Forgetting of Imagenet1k and various finetuned classification datasets, after fine-tuning. Maximum mean and standard deviation of 4 independent runs, among evaluations computed from epoch 50. Ablation with DINOv2 backbone.}
    \begin{tabular}{cccccccc}
    \toprule
    Methods & \multicolumn{2}{c}{CIFAR10} & \multicolumn{2}{c}{CIFAR100} & \multicolumn{2}{c}{DTD} \\
     & Acc & Forg & Acc & Forg & Acc & Forg \\
    \midrule
    LoRA & $\bm{99.61}_{\pm0.03}$ & $5.34_{\pm1.31}$ & $95.40_{\pm0.16}$ & $14.44_{\pm1.85}$ & $\bm{86.38}_{\pm0.45}$ & $28.65_{\pm2.28}$ \\
    DoRA & $\bm{99.61}_{\pm0.01}$ & $5.69_{\pm1.13}$ & $95.38_{\pm0.09}$ & $13.91_{\pm1.63}$ & $86.26_{\pm0.63}$ & $30.18_{\pm10.60}$ \\
    OLoRA & $99.57_{\pm0.05}$ & $10.83_{\pm3.32}$ & $95.22_{\pm0.07}$ & $13.89_{\pm0.49}$ & $85.66_{\pm0.21}$ & $17.93_{\pm2.32}$ \\
    \midrule
    0 (PiSSA) & $99.48_{\pm0.03}$ & $4.36_{\pm0.25}$ & $94.96_{\pm0.05}$ & $12.63_{\pm0.78}$ & $85.51_{\pm0.52}$ & $19.46_{\pm4.99}$ \\
    \rowcolor{yellow!20}256 (25\%) & $99.60_{\pm0.05}$ & $2.63_{\pm0.38}$ & $95.37_{\pm0.10}$ & $\bm{3.00}_{\pm0.66}$ & $86.00_{\pm0.61}$ & $\bm{10.78}_{\pm1.46}$ \\
    \rowcolor{yellow!20}512 (50\%) & $\bm{99.61}_{\pm0.03}$ & $\bm{2.34}_{\pm0.13}$ & $95.32_{\pm0.11}$ & $3.37_{\pm0.57}$ & $86.04_{\pm0.39}$ & $14.16_{\pm2.18}$ \\
    \rowcolor{yellow!20}768 (75\%) & $\bm{99.61}_{\pm0.03}$ & $2.39_{\pm0.52}$ & $\bm{95.43}_{\pm0.13}$ & $6.51_{\pm0.50}$ & $86.16_{\pm0.73}$ & $16.60_{\pm4.91}$ \\
    992 (MiLoRA) & $99.59_{\pm0.03}$ & $3.81_{\pm0.55}$ & $95.30_{\pm0.08}$ & $5.40_{\pm0.71}$ & $86.04_{\pm0.27}$ & $31.05_{\pm8.15}$ \\
    \bottomrule
    \end{tabular}
    
    \begin{tabular}{cccccccc}
    \toprule
    Methods & \multicolumn{2}{c}{Caltech101} & \multicolumn{2}{c}{Caltech256} & \multicolumn{2}{c}{Food101} \\
     & Acc & Forg & Acc & Forg & Acc & Forg \\
    \midrule
    LoRA & $93.39_{\pm0.71}$ & $10.68_{\pm3.31}$ & $97.04_{\pm0.18}$ & $9.92_{\pm1.42}$ & $95.35_{\pm0.09}$ & $3.29_{\pm0.33}$ \\
    DoRA & $\bm{93.49}_{\pm0.94}$ & $18.22_{\pm2.57}$ & $96.95_{\pm0.28}$ & $10.54_{\pm0.77}$ & $95.35_{\pm0.05}$ & $3.46_{\pm0.35}$ \\
    OLoRA & $93.33_{\pm0.72}$ & $12.85_{\pm1.87}$ & $96.95_{\pm0.10}$ & $10.50_{\pm0.75}$ & $95.10_{\pm0.03}$ & $7.34_{\pm0.50}$ \\
    \midrule
    0 (PiSSA) & $92.64_{\pm0.66}$ & $10.13_{\pm0.47}$ & $96.63_{\pm0.13}$ & $11.21_{\pm0.28}$ & $95.02_{\pm0.03}$ & $5.85_{\pm0.34}$ \\
    \rowcolor{yellow!20}256 (25\%) & $92.82_{\pm0.53}$ & $7.19_{\pm0.96}$ & $97.12_{\pm0.28}$ & $7.68_{\pm0.35}$ & $95.36_{\pm0.13}$ & $2.81_{\pm0.20}$ \\
    \rowcolor{yellow!20}512 (50\%) & $92.64_{\pm0.59}$ & $\bm{6.91}_{\pm1.10}$ & $97.07_{\pm0.11}$ & $7.61_{\pm0.86}$ & $\bm{95.43}_{\pm0.08}$ & $2.45_{\pm0.25}$ \\
    \rowcolor{yellow!20}768 (75\%) & $93.35_{\pm0.43}$ & $7.23_{\pm1.09}$ & $97.00_{\pm0.15}$ & $\bm{7.60}_{\pm1.16}$ & $95.35_{\pm0.03}$ & $\bm{2.44}_{\pm0.20}$ \\
    992 (MiLoRA) & $93.39_{\pm0.40}$ & $12.20_{\pm2.14}$ & $\bm{97.13}_{\pm0.15}$ & $8.55_{\pm1.03}$ & $95.36_{\pm0.06}$ & $2.73_{\pm0.15}$ \\
    \bottomrule
    \end{tabular}
    
    \begin{tabular}{cccccccc}
    \toprule
    Methods & \multicolumn{2}{c}{O. Pets} & \multicolumn{2}{c}{O. Flowers} & \multicolumn{2}{c}{S. Cars} \\
     & Acc & Forg & Acc & Forg & Acc & Forg \\
    \midrule
    LoRA & $96.33_{\pm0.15}$ & $5.45_{\pm0.43}$ & $99.70_{\pm0.02}$ & $3.42_{\pm0.56}$ & $94.42_{\pm0.28}$ & $4.27_{\pm0.28}$ \\
    DoRA & $96.27_{\pm0.30}$ & $5.74_{\pm0.53}$ & $99.69_{\pm0.02}$ & $3.52_{\pm0.61}$ & $94.40_{\pm0.23}$ & $4.31_{\pm0.23}$ \\
    OLoRA & $96.07_{\pm0.16}$ & $8.43_{\pm1.27}$ & $99.68_{\pm0.09}$ & $7.30_{\pm0.86}$ & $94.31_{\pm0.12}$ & $11.90_{\pm1.59}$ \\
    \midrule
    0 (PiSSA) & $95.91_{\pm0.05}$ & $7.37_{\pm1.27}$ & $99.67_{\pm0.02}$ & $4.99_{\pm0.52}$ & $94.20_{\pm0.26}$ & $7.98_{\pm0.82}$ \\
    \rowcolor{yellow!20}256 (25\%) & $96.33_{\pm0.09}$ & $3.89_{\pm0.41}$ & $99.68_{\pm0.02}$ & $\bm{2.33}_{\pm0.14}$ & $94.51_{\pm0.18}$ & $3.25_{\pm0.59}$ \\
    \rowcolor{yellow!20}512 (50\%) & $96.39_{\pm0.24}$ & $\bm{3.59}_{\pm0.46}$ & $99.69_{\pm0.03}$ & $2.36_{\pm0.10}$ & $94.47_{\pm0.15}$ & $\bm{3.09}_{\pm0.28}$ \\
    \rowcolor{yellow!20}768 (75\%) & $\bm{96.41}_{\pm0.18}$ & $3.69_{\pm0.31}$ & $\bm{99.71}_{\pm0.01}$ & $\bm{2.33}_{\pm0.23}$ & $94.43_{\pm0.03}$ & $3.26_{\pm0.15}$ \\
    992 (MiLoRA) & $96.38_{\pm0.20}$ & $5.10_{\pm0.78}$ & $99.70_{\pm0.02}$ & $3.20_{\pm0.30}$ & $\bm{94.55}_{\pm0.07}$ & $3.64_{\pm0.22}$ \\
    \bottomrule
    \end{tabular}
    
    \begin{tabular}{cccccccc}
    \toprule
    Methods & \multicolumn{2}{c}{S. Dogs} & \multicolumn{2}{c}{FGVC Aircraft} & \multicolumn{2}{c}{Average} \\
     & Acc & Forg & Acc & Forg & Acc & Forg \\
    \midrule
    LoRA & $89.94_{\pm0.24}$ & $8.37_{\pm1.61}$ & $94.73_{\pm0.31}$ & $1.95_{\pm0.18}$ & $94.75_{\pm0.24}$ & $8.71_{\pm1.23}$ \\
    DoRA & $89.91_{\pm0.12}$ & $8.51_{\pm1.27}$ & $94.58_{\pm0.12}$ & $1.92_{\pm0.15}$ & $94.72_{\pm0.25}$ & $9.64_{\pm1.80}$ \\
    OLoRA & $89.77_{\pm0.22}$ & $10.57_{\pm1.44}$ & $94.49_{\pm0.36}$ & $11.81_{\pm1.78}$ & $94.56_{\pm0.19}$ & $11.21_{\pm1.47}$ \\
    \midrule
    0 (PiSSA) & $89.55_{\pm0.28}$ & $9.99_{\pm1.62}$ & $\bm{94.78}_{\pm0.48}$ & $7.68_{\pm1.21}$ & $94.40_{\pm0.23}$ & $9.24_{\pm1.14}$ \\
    \rowcolor{yellow!20}256 (25\%) & $90.12_{\pm0.21}$ & $4.86_{\pm0.55}$ & $94.60_{\pm0.32}$ & $1.61_{\pm0.07}$ & $94.68_{\pm0.23}$ & $\bm{4.55}_{\pm0.52}$ \\
    \rowcolor{yellow!20}512 (50\%) & $90.13_{\pm0.13}$ & $\bm{4.41}_{\pm0.44}$ & $94.70_{\pm0.23}$ & $\bm{1.43}_{\pm0.06}$ & $94.68_{\pm0.19}$ & $4.70_{\pm0.58}$ \\
    \rowcolor{yellow!20}768 (75\%) & $\bm{90.22}_{\pm0.22}$ & $4.75_{\pm0.63}$ & $94.73_{\pm0.33}$ & $1.47_{\pm0.18}$ & $\bm{94.76}_{\pm0.21}$ & $5.30_{\pm0.90}$ \\
    992 (MiLoRA) & $90.17_{\pm0.16}$ & $5.85_{\pm0.72}$ & $94.22_{\pm0.37}$ & $2.09_{\pm0.35}$ & $94.71_{\pm0.16}$ & $7.60_{\pm1.37}$ \\
    \bottomrule
    \end{tabular}
    \label{tab:cls_dinov2}
\end{table*}

\newpage
\clearpage
\subsection{NLP}

\begin{figure*}[h]
    \centering

    \begin{subfigure}{0.32\textwidth}
        \centering
        \includegraphics[width=\textwidth]{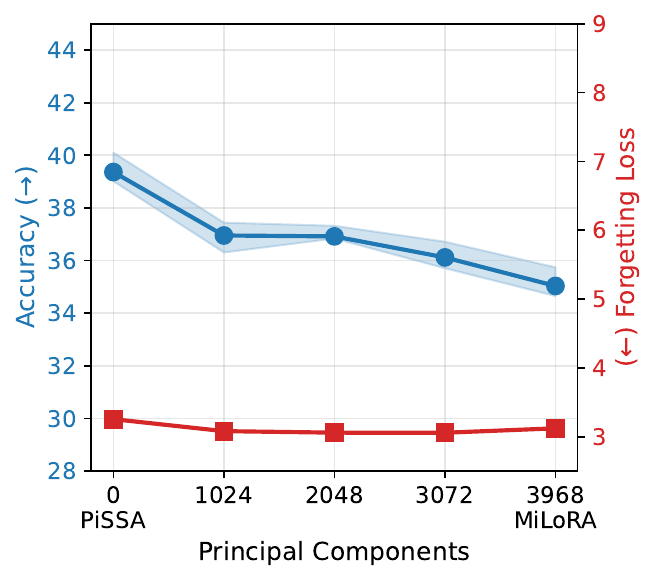}
        \caption{PISSA setup.}
    \end{subfigure}
    \hfill
    \begin{subfigure}{0.32\textwidth}
        \centering
        \includegraphics[width=\textwidth]{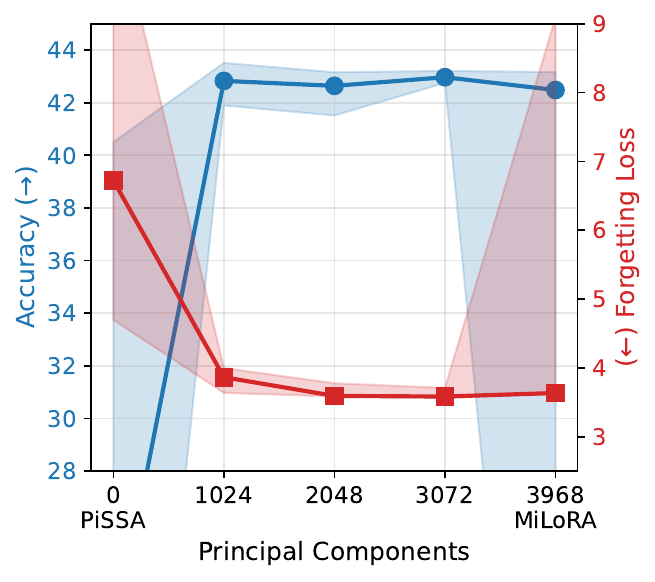}
        \caption{Ext setup with  lr: 3e-4.}
    \end{subfigure}
    \hfill
    \begin{subfigure}{0.32\textwidth}
        \centering
        \includegraphics[width=\textwidth]{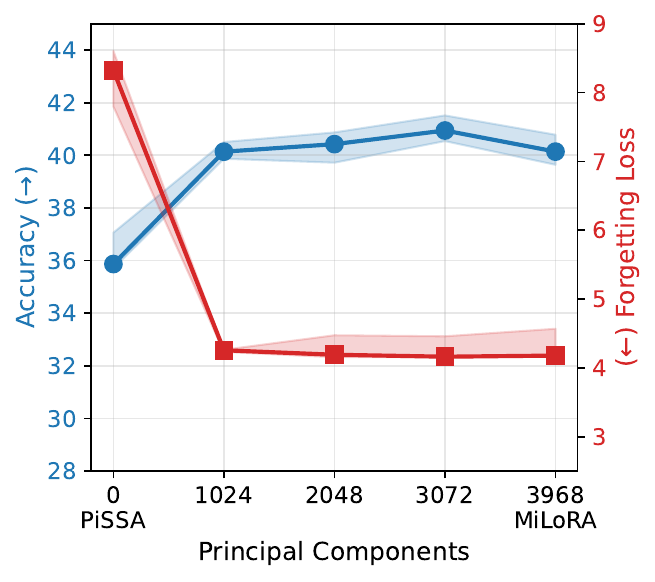}
        \caption{MiLoRA setup.}
    \end{subfigure}

    \begin{subfigure}{0.32\textwidth}
        \centering
        \includegraphics[width=\textwidth]{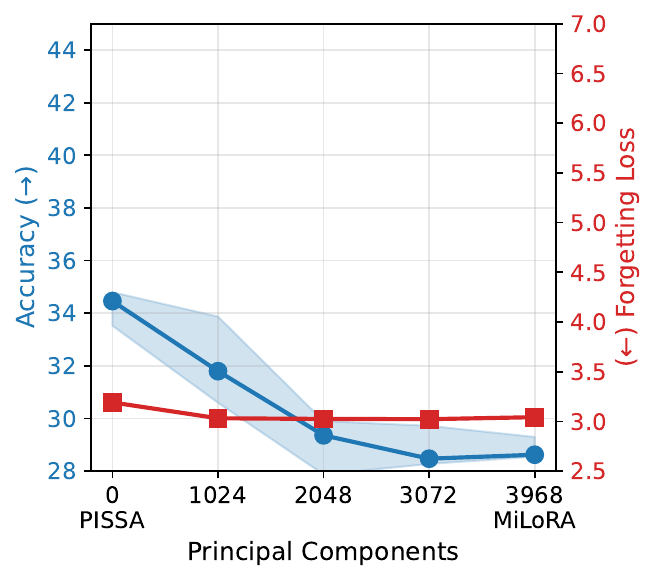}
        \caption{PISSA setup.}
    \end{subfigure}
    \hfill
    \begin{subfigure}{0.32\textwidth}
        \centering
        \includegraphics[width=\textwidth]{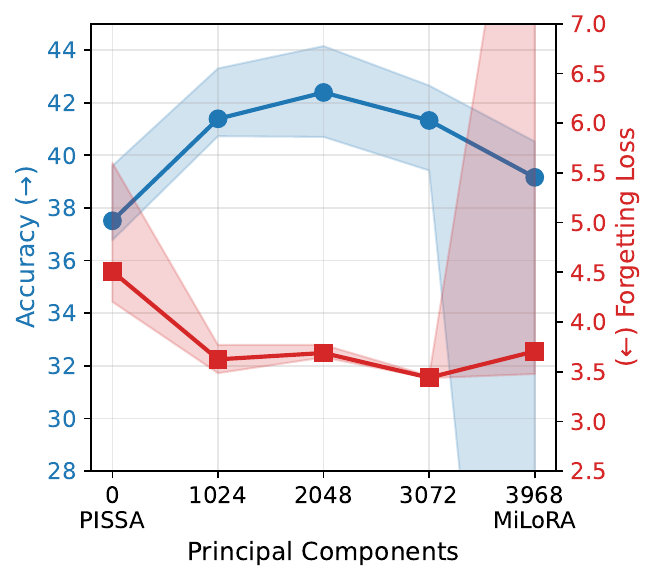}
        \caption{Ext setup with  lr: 3.5e-4.}
    \end{subfigure}
    \hfill
    \begin{subfigure}{0.32\textwidth}
        \centering
        \includegraphics[width=\textwidth]{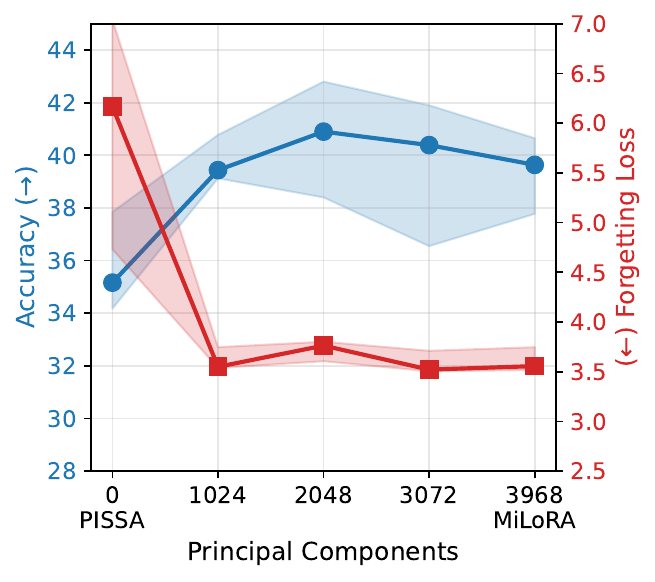}
        \caption{MiLoRA setup.}
    \end{subfigure}

  \begin{subfigure}[b]{0.32\textwidth}
        \centering
        \includegraphics[width=\textwidth]{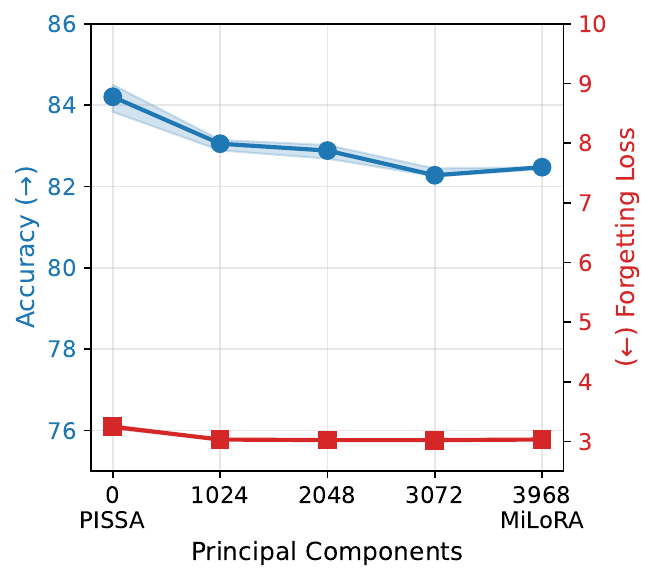}
        \caption{PISSA setup.}
    \end{subfigure}
    \hfill
    \begin{subfigure}[b]{0.32\textwidth}
        \centering
        \includegraphics[width=\textwidth]{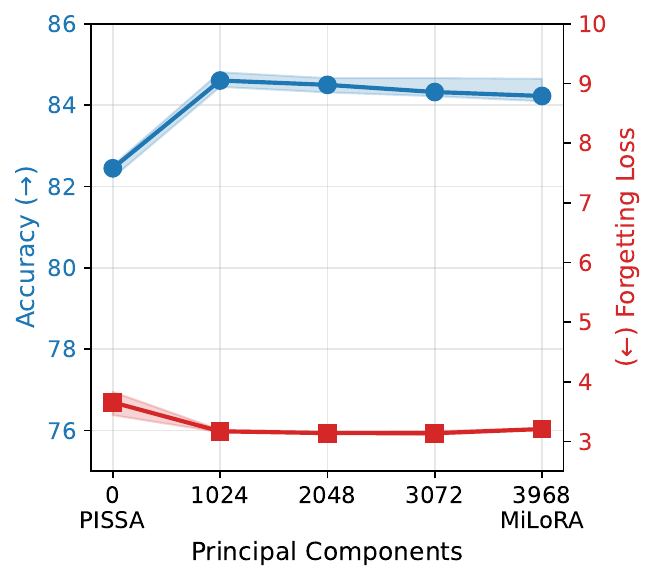}
        \caption{Ext setup with  lr: 1e-4.}
    \end{subfigure}
    \hfill
    \begin{subfigure}[b]{0.32\textwidth}
        \centering
        \includegraphics[width=\textwidth]{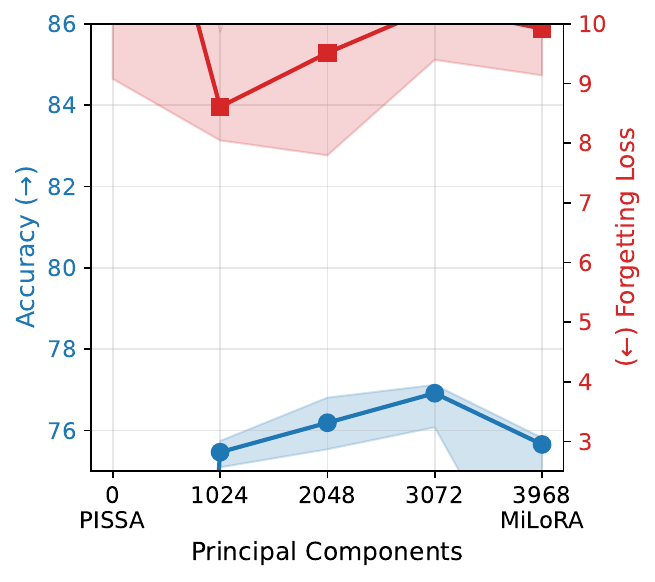}
        \caption{MiLoRA setup.}
    \end{subfigure}
    
    \caption{Mathematical reasoning (top row), Python coding (middle row) and Common sense (bottom row) with LLaMA-2 7b. We report median and min/max. Outlier values are runs with exploding gradients.}
    \label{fig:llm}
\end{figure*}

\clearpage
\newpage

\begin{table*}[t]
    \centering
    \small
    \caption{Mathematical reasoning results with LLaMA-2 7b. We report mean and standard deviation over 4 independent runs. High standard deviations include runs with exploding gradients. We highlight $\textbf{best}$,  $\underline{\text{second best}}$ and in \textcolor{yellow}{yellow} the best rows overall.}
    \begin{tabular}{cccccc}
    \toprule
    Methods & Setup & MATH & GSM8K & Average ($\uparrow$) & Forgetting ($\downarrow$) \\
    \midrule
    DoRA & ext & $15.35_{\pm0.291}$ & $63.80_{\pm0.85}$ & $39.57_{\pm0.48}$ & $3.29_{\pm0.01}$ \\
    OLoRA & ext & $14.87_{\pm0.35}$ & $61.24_{\pm1.15}$ & $38.05_{\pm0.53}$ & $4.31_{\pm0.03}$ \\
    \midrule
    PiSSA & pissa & $14.33_{\pm0.30}$ & $64.59_{\pm0.73}$ & $39.46_{\pm0.50}$ & $\bm{3.26}_{\pm0.00}$ \\
    PiSSA & ext & $9.79_{\pm9.82}$ & $32.75_{\pm34.50}$ & $21.27_{\pm22.16}$ & $7.23_{\pm3.03}$ \\
    \rowcolor{yellow!20}1024 (25\%) & ext & $\bm{19.59}_{\pm0.14}$ & $65.94_{\pm1.24}$ & $42.77_{\pm0.67}$ & $3.84_{\pm0.16}$ \\
    \rowcolor{yellow!20}2048 (50\%) & ext & $18.82_{\pm0.39}$ & $\underline{66.15}_{\pm1.03}$ & $\underline{42.49}_{\pm0.69}$ & $3.64_{\pm0.09}$ \\
    \rowcolor{yellow!20}3072 (75\%) & ext & $\underline{19.43}_{\pm0.16}$ & $\bm{66.51}_{\pm0.44}$ & $\bm{42.97}_{\pm0.25}$ & $\underline{3.61}_{\pm0.07}$ \\
    MiLoRA & ext & $14.02_{\pm9.36}$ & $50.04_{\pm33.36}$ & $32.03_{\pm21.36}$ & $5.00_{\pm2.73}$ \\
    MiLoRA & milora & $17.35_{\pm0.62}$ & $63.00_{\pm0.48}$ & $40.18_{\pm0.47}$ & $4.27_{\pm0.20}$ \\
    \bottomrule
    \end{tabular}
    \label{tab:llm_metamath}
\end{table*}

\begin{table*}[t]
    \centering
    \scriptsize
    \caption{Python coding results with LLaMA-2 7b. We report mean and standard deviation over 4 independent runs. High standard deviations include runs with exploding gradients. We highlight $\textbf{best}$, $\underline{\text{second best}}$ and in \textcolor{yellow}{yellow} the best rows overall.}
    \begin{tabular}{cccccccc}
    \toprule
    Methods & Setup & Human-eval & Human-eval+ & MBPP & MBPP+ & Average ($\uparrow$) & Forgetting ($\downarrow$) \\
    \midrule
    DoRA & ext & $35.68_{\pm2.26}$ & $32.15_{\pm3.15}$ & $40.15_{\pm2.92}$ & $33.88_{\pm1.76}$ & $35.46_{\pm2.07}$ & $3.13_{\pm0.00}$ \\
    OLoRA & ext & $33.05_{\pm1.02}$ & $28.82_{\pm0.62}$ & $37.22_{\pm0.88}$ & $31.28_{\pm2.26}$ & $32.59_{\pm1.09}$ & $3.61_{\pm0.02}$ \\
    \midrule
    0 (PiSSA) & pissa & $31.23_{\pm2.36}$ & $28.98_{\pm2.09}$ & $42.25_{\pm1.71}$ & $34.80_{\pm1.65}$ & $34.31_{\pm0.56}$ & $\bm{3.19}_{\pm0.00}$ \\
    0 (PiSSA) & ext & $38.73_{\pm1.91}$ & $35.23_{\pm2.02}$ & $42.08_{\pm1.08}$ & $35.38_{\pm1.16}$ & $37.85_{\pm1.35}$ & $4.71_{\pm0.64}$ \\
    \rowcolor{yellow!20}1024 (25\%) & ext & $\underline{43.92}_{\pm2.13}$ & $\underline{39.32}_{\pm2.09}$ & $44.65_{\pm0.67}$ & $\bm{38.90}_{\pm0.48}$ & $\underline{41.70}_{\pm1.17}$ & $3.63_{\pm0.12}$ \\
    \rowcolor{yellow!20}2048 (50\%) & ext & $\bm{44.22}_{\pm4.03}$ & $\bm{40.10}_{\pm4.17}$ & $\bm{46.48}_{\pm1.24}$ & $\underline{38.82}_{\pm1.07}$ & $\bm{42.41}_{\pm1.65}$ & $3.70_{\pm0.06}$ \\
    \rowcolor{yellow!20}3072 (50\%) & ext & $43.30_{\pm4.02}$ & $38.12_{\pm3.34}$ & $\underline{45.67}_{\pm1.92}$ & $37.62_{\pm0.69}$ & $41.18_{\pm1.37}$ & $\underline{3.44}_{\pm0.01}$ \\
    3968 (MiLoRA) & ext & $30.97_{\pm20.65}$ & $27.45_{\pm18.32}$ & $32.73_{\pm21.92}$ & $27.70_{\pm18.47}$ & $29.71_{\pm19.83}$ & $5.25_{\pm3.25}$ \\
    3968 (MiLoRA) & milora & $39.02_{\pm1.81}$ & $35.83_{\pm1.77}$ & $45.17_{\pm1.50}$ & $37.68_{\pm1.76}$ & $39.42_{\pm1.40}$ & $3.59_{\pm0.10}$ \\
    \bottomrule
    \end{tabular}
    \label{tab:llm_python}
\end{table*}

\begin{table*}[h]
    \centering
    \scriptsize
    \caption{Common sense results with LLaMA-2 7b.  We report mean and standard deviation over 4 independent runs. We highlight $\textbf{best}$,  $\underline{\text{second best}}$ and in \textcolor{yellow}{yellow} the best rows overall.}
    \begin{tabular}{cccccccc}
    \toprule
    Methods & Setup & BoolQ & PIQA & SIQA & HellaSwag & WinoGrande & ARC-e \\
    \midrule
    DoRA & ext & $73.13_{\pm0.38}$ & $85.65_{\pm0.48}$ & $80.89_{\pm0.62}$ & $94.37_{\pm0.09}$ & $85.00_{\pm0.57}$ & $87.97_{\pm0.50}$ \\
    OLoRA & ext & $73.72_{\pm0.42}$ & $84.98_{\pm0.40}$ & $80.49_{\pm0.64}$ & $93.85_{\pm0.13}$ & $85.58_{\pm0.62}$ & $87.30_{\pm0.65}$ \\
    \midrule
    PiSSA & pissa & $74.62_{\pm0.55}$ & $86.00_{\pm0.46}$ & $\underline{81.53}_{\pm0.54}$ & $94.57_{\pm0.26}$ & $86.78_{\pm0.28}$ & $88.86_{\pm0.49}$ \\
    PiSSA & ext & $73.06_{\pm0.52}$ & $84.33_{\pm0.63}$ & $80.62_{\pm0.49}$ & $92.67_{\pm0.28}$ & $84.87_{\pm0.78}$ & $86.09_{\pm0.92}$ \\
    \rowcolor{yellow!20}1024 (25\%) & ext & $\bm{75.04}_{\pm0.40}$ & $\bm{86.24}_{\pm0.69}$ & $81.36_{\pm0.40}$ & $\bm{95.06}_{\pm0.06}$ & $86.90_{\pm0.39}$ & $\bm{89.19}_{\pm0.57}$ \\
    \rowcolor{yellow!20}2048 (50\%) & ext & $\underline{74.68}_{\pm0.62}$ & $85.89_{\pm0.17}$ & $81.47_{\pm0.19}$ & $\underline{94.97}_{\pm0.12}$ & $86.44_{\pm0.24}$ & $88.88_{\pm0.58}$ \\
    \rowcolor{yellow!20}3072 (75\%) & ext & $74.54_{\pm0.14}$ & $\underline{86.15}_{\pm0.40}$ & $81.13_{\pm0.44}$ & $\underline{94.97}_{\pm0.17}$ & $\underline{87.10}_{\pm0.59}$ & $88.82_{\pm0.29}$ \\
    3968 (MiLoRA) & ext & $74.27_{\pm0.52}$ & $86.10_{\pm0.46}$ & $\bm{81.61}_{\pm0.14}$ & $94.91_{\pm0.33}$ & $\bm{87.25}_{\pm0.56}$ & $\underline{89.07}_{\pm0.33}$ \\
    3968 (MiLoRA) & milora & $69.72_{\pm1.33}$ & $77.77_{\pm1.96}$ & $75.67_{\pm1.20}$ & $83.50_{\pm4.01}$ & $77.05_{\pm1.92}$ & $76.76_{\pm2.75}$ \\
    \bottomrule
    \end{tabular}
    
    \begin{tabular}{cccccc}
    \toprule
    Methods & Setup & ARC-c & OBQA & Average ($\uparrow$) & Forgetting ($\downarrow$) \\
    \midrule
    DoRA & ext & $75.09_{\pm0.52}$ & $84.15_{\pm0.41}$ & $83.28_{\pm0.21}$ & $\bm{3.03}_{\pm0.00}$ \\
    OLoRA & ext & $73.81_{\pm0.63}$ & $85.50_{\pm0.38}$ & $83.15_{\pm0.21}$ & $3.83_{\pm0.08}$ \\
    \midrule
    0 (PiSSA) & pissa & $75.83_{\pm0.75}$ & $85.28_{\pm1.16}$ & $84.19_{\pm0.27}$ & $3.24_{\pm0.04}$ \\
    0 (PiSSA) & ext & $73.02_{\pm0.34}$ & $84.60_{\pm1.14}$ & $82.41_{\pm0.13}$ & $3.64_{\pm0.20}$ \\
    \rowcolor{yellow!20}1024 (25\%) & ext & $76.09_{\pm0.43}$ & $\bm{87.05}_{\pm0.41}$ & $\bm{84.62}_{\pm0.17}$ & $3.17_{\pm0.02}$ \\
    \rowcolor{yellow!20}2048 (50\%) & ext & $\bm{77.05}_{\pm0.68}$ & $\underline{86.55}_{\pm0.77}$ & $\underline{84.49}_{\pm0.19}$ & $\underline{3.14}_{\pm0.01}$ \\
    \rowcolor{yellow!20}3072 (75\%) &ext & $\underline{76.54}_{\pm0.21}$ & $85.80_{\pm1.10}$ & $84.38_{\pm0.20}$ & $\underline{3.14}_{\pm0.03}$ \\
    3968 (MiLoRA) & ext & $75.96_{\pm0.34}$ & $85.20_{\pm0.71}$ & $84.30_{\pm0.25}$ & $3.21_{\pm0.01}$ \\
    3968 (MiLoRA) & milora & $61.35_{\pm2.70}$ & $75.20_{\pm2.47}$ & $74.63_{\pm2.17}$ & $10.68_{\pm2.14}$ \\
    \bottomrule
    \end{tabular}
    \label{tab:llm_commonsense}
\end{table*}



\end{document}